\documentclass{article}

\usepackage[table]{xcolor} 

\usepackage[preprint]{corl_2026} 

\definecolor{Light1}{rgb}{0.98, 0.95, 0.90}
\definecolor{Light2}{rgb}{0.98, 0.98, 0.93}
\definecolor{Light3}{rgb}{0.98, 0.98, 1}
\definecolor{Light4}{rgb}{0.93, 0.98, 0.98}

\usepackage{graphicx} 
\usepackage{mathtools}
\usepackage{bm} 
\usepackage{xspace} 

\usepackage{booktabs}

\usepackage{listings}

\usepackage{algorithm}
\usepackage{algorithmicx}
\usepackage{algpseudocode}
\makeatletter
\renewcommand{\theHALG@line}{\thealgorithm.\theALG@line}
\makeatother

\definecolor{profblue}{HTML}{1f77b4}       
\definecolor{proforange}{HTML}{ff7f0e}     
\definecolor{profgreen}{HTML}{A2AD00}      
\definecolor{profdarkgray}{HTML}{333333}   

\definecolor{pyblue}{RGB}{0,102,204}      
\definecolor{pygreen}{RGB}{0,128,0}       
\definecolor{pyred}{RGB}{187,0,0}         
\definecolor{pypurple}{RGB}{128,0,128}    
\definecolor{pyorange}{RGB}{255,102,0}    
\definecolor{pygray}{RGB}{128,128,128}    
\definecolor{pybackground}{RGB}{248,248,248} 

\usepackage{makecell}
\usepackage{multirow}

\usepackage[font=footnotesize]{caption}
\usepackage{subcaption}

\usepackage{wrapfig}

\usepackage[capitalize]{cleveref}
\Crefname{section}{Section}{Sections}
\Crefname{table}{Table}{Tables}

\usepackage{threeparttable}

\algnewcommand{\LeftComment}[1]{\Statex \(\triangleright\) #1}

\usepackage{amsmath,amssymb,amsfonts}
\def\BibTeX{{\rm B\kern-.05em{\sc i\kern-.025em b}\kern-.08em
    T\kern-.1667em\lower.7ex\hbox{E}\kern-.125emX}}

\Crefname{equation}{Eq.}{Eqs.} 
\Crefname{figure}{Fig.}{Figs.}

\graphicspath{{images-src/}{}} 

\definecolor{dlrprim1}{HTML}{000000} 
\definecolor{dlrprim2}{HTML}{666666} 
\definecolor{dlrprim3}{HTML}{b9cad2}
\definecolor{dlrprim4}{HTML}{ffffff} 

\definecolor{dlrblue1}{HTML}{00658b} 
\definecolor{dlrblue2}{HTML}{3b98cb}
\definecolor{dlrblue3}{HTML}{6cb9dc}
\definecolor{dlrblue4}{HTML}{a7d3ec}
\definecolor{dlrblue5}{HTML}{d1e8fa}

\definecolor{dlryellow1}{HTML}{d2ae3d}
\definecolor{dlryellow2}{HTML}{f2cd51}
\definecolor{dlryellow3}{HTML}{f8de53}
\definecolor{dlryellow4}{HTML}{fcea7a}
\definecolor{dlryellow5}{HTML}{fff8be}

\definecolor{dlrgreen1}{HTML}{82a043} 
\definecolor{dlrgreen2}{HTML}{a6bf51}
\definecolor{dlrgreen3}{HTML}{cad55c}
\definecolor{dlrgreen4}{HTML}{d9df78}
\definecolor{dlrgreen5}{HTML}{e6eaaf}

\definecolor{dlrgray1}{HTML}{666666} 
\definecolor{dlrgray2}{HTML}{868585}
\definecolor{dlrgray3}{HTML}{b1b1b1}
\definecolor{dlrgray4}{HTML}{cfcfcf}
\definecolor{dlrgray5}{HTML}{ebebeb}

\usepackage[textsize=scriptsize]{todonotes} 
\tikzset{
	cross/.pic = {
		\draw[rotate = 45, line width=0.75pt] (-#1,0) -- (#1,0);
		\draw[rotate = 45, line width=0.75pt] (0,-#1) -- (0, #1);
	}
}

\newcommand{\annotategraphicsmulti}[3][]{
  \begin{tikzpicture}[%
  every node/.style={draw=black, black, text opacity=1, fill=white, fill opacity=0.75,inner sep=0.5mm, #1},%
  ]
  \node[anchor=south west,inner sep=0, draw=none] (image) at (0,0) {
    #2
  };
  \begin{scope}[x={(image.south east)},y={(image.north west)}]
    #3
  \end{scope}
  \end{tikzpicture}%
}

\newcommand{\eg}{e.g.\@\xspace}  
\newcommand{\ie}{i.e.\@\xspace}  

\usepackage{orcidlink}

\newcommand{\timeStep}{t}

\newcommand{\mean}{\bm{\mu}}
\newcommand{\covariance}{\bm{\Sigma}}

\newcommand{\inputVariable}{\bm{s}}
\newcommand{\outputVariable}{\bm{\xi}}

\newcommand{\inputDimension}{\mathcal{I}}
\newcommand{\outputDimension}{\mathcal{O}}

\newcommand{\local}{(p)}
\newcommand{\localMean}{\mean^{\local}}
\newcommand{\localCovariance}{\covariance^{\local}}
\newcommand{\amountOfFrames}{P}
\newcommand{\frameIndex}{p}

\newcommand{\frameOrigin}{\bm{b}}
\newcommand{\localFrameOrigin}{\frameOrigin^{\local}}
\newcommand{\frameOriginAtTimestep}{\frameOrigin_{\timeStep}}
\newcommand{\frameRotationMatrix}{\bm{A}}
\newcommand{\localFrameRotationMatrix}{\frameRotationMatrix^{\local}}
\newcommand{\frameRotationMatrixAtTimestep}{\bm{A}_{\timeStep}}

\newcommand{\amountOfDatapoints}{M}
\newcommand{\datapointIndex}{m}

\newcommand{\trajectoryLength}{H}
\newcommand{\trajectoryIndex}{h}

\newcommand{\amountOfKMP}{N}
\newcommand{\kmpIndex}{n}

\newcommand{\kmp}{\bm{D}}
\newcommand{\kmpInput}{\inputVariable_{\kmpIndex}}

\newcommand{\kmpMean}{\mean_{\kmpIndex}}
\newcommand{\localkmpMean}{\kmpMean^{\local}}
\newcommand{\kmpCovariance}{\covariance_{\kmpIndex}}
\newcommand{\localkmpCovariance}{\kmpCovariance^{\local}}
\newcommand{\localkmp}{\Theta^{\local}}

\newcommand{\tpkmp}{\Theta}

\newcommand{\force}{\bm{F}}

\newcommand{\orcidMarkus}{0000-0001-8229-9410}

\newcommand{\orcidAlin}{0000-0001-5343-9074}
\newcommand{\orcidFreek}{0000-0001-9555-9517}
\newcommand{\orcidJoao}{0000-0003-1428-8933}
\newcommand{\orcidSamuel}{0000-0002-7923-8307}

\newcommand{\orcidValentin}{0009-0004-3715-334X}
\newcommand{\orcidTai}{0009-0008-5354-8666}
\newcommand{\addorcidMarkus}{\orcidlink{\orcidMarkus}}
\newcommand{\addorcidAlin}{\orcidlink{\orcidAlin}}
\newcommand{\addorcidFreek}{\orcidlink{\orcidFreek}}
\newcommand{\addorcidJoao}{\orcidlink{\orcidJoao}}
\newcommand{\addorcidSamuel}{\orcidlink{\orcidSamuel}}

\newcommand{\addorcidValentin}{\orcidlink{\orcidValentin}}
\newcommand{\addorcidTai}{\orcidlink{\orcidTai}}

\begin{document}

\title{
	CLASP: Language-Driven Robot Skill Selection and Composition using Task-Parameterized Learning
} 

\author{
	Markus Knauer\addorcidMarkus$^{1, 2}$
	\And Valentin Gieraths\addorcidValentin$^{1, 2}$
	\And Tai Mai\addorcidTai$^{1}$
	\And Samuel Bustamante\addorcidSamuel$^{1, 2}$
	\AND
	Alin Albu-Sch\"affer\addorcidAlin$^{1, 2}$
	\And Freek Stulp\addorcidFreek$^{1}$
	\And Jo\~ao Silv\'erio\addorcidJoao$^{1}$
	\AND
	\normalfont\normalsize
	$^{1}$German Aerospace Center (DLR), Institute of Robotics and Mechatronics (RMC), \\
	M\"unchener Str. 20, 82234 We\ss ling, Germany.\quad\texttt{\{first\}.\{last\}@dlr.de} \\[0.3em]
	$^{2}$School of Computation, Information and Technology (CIT), \\
	Technical University of Munich (TUM), Arcisstr.~21, 80333 Munich, Germany.\quad\texttt{m.knauer@tum.de}
}


\maketitle

\begin{abstract}
	Enabling robots to understand and execute tasks from natural language commands while maintaining data efficiency remains challenging.
	Foundation models such as vision-language-action (VLA) and vision-language models (VLMs) provide intuitive interaction channels but require extensive data; task-parameterized imitation learning achieves data efficiency but lacks natural language grounding.
	This work bridges this gap through a modular architecture combining task-parameterized kernelized movement primitives (TP-KMPs) with pretrained VLMs.
	\textit{During learning}, skills are acquired from 2--5 kinesthetic demonstrations, and the VLM generates skill schemas describing each skill's parameters and preconditions.
	\textit{During execution}, the VLM interprets commands to select skills, reason about parameter bindings, and create novel behaviors through covariance-weighted composition.
	\textit{When no skill or composition suffices}, the system identifies capability gaps and requests targeted demonstrations, all without fine-tuning.
	Validation on a 7-DoF manipulator shows success rates of 73.3\%--100\% in scenarios requiring skill selection, composition, and active learning.
\end{abstract}





\section{Introduction}
\label{sec:introduction}
\begin{figure}[!tb]
	\centering
	\raisebox{-0.5\height}{\includegraphics[width=0.60\textwidth]{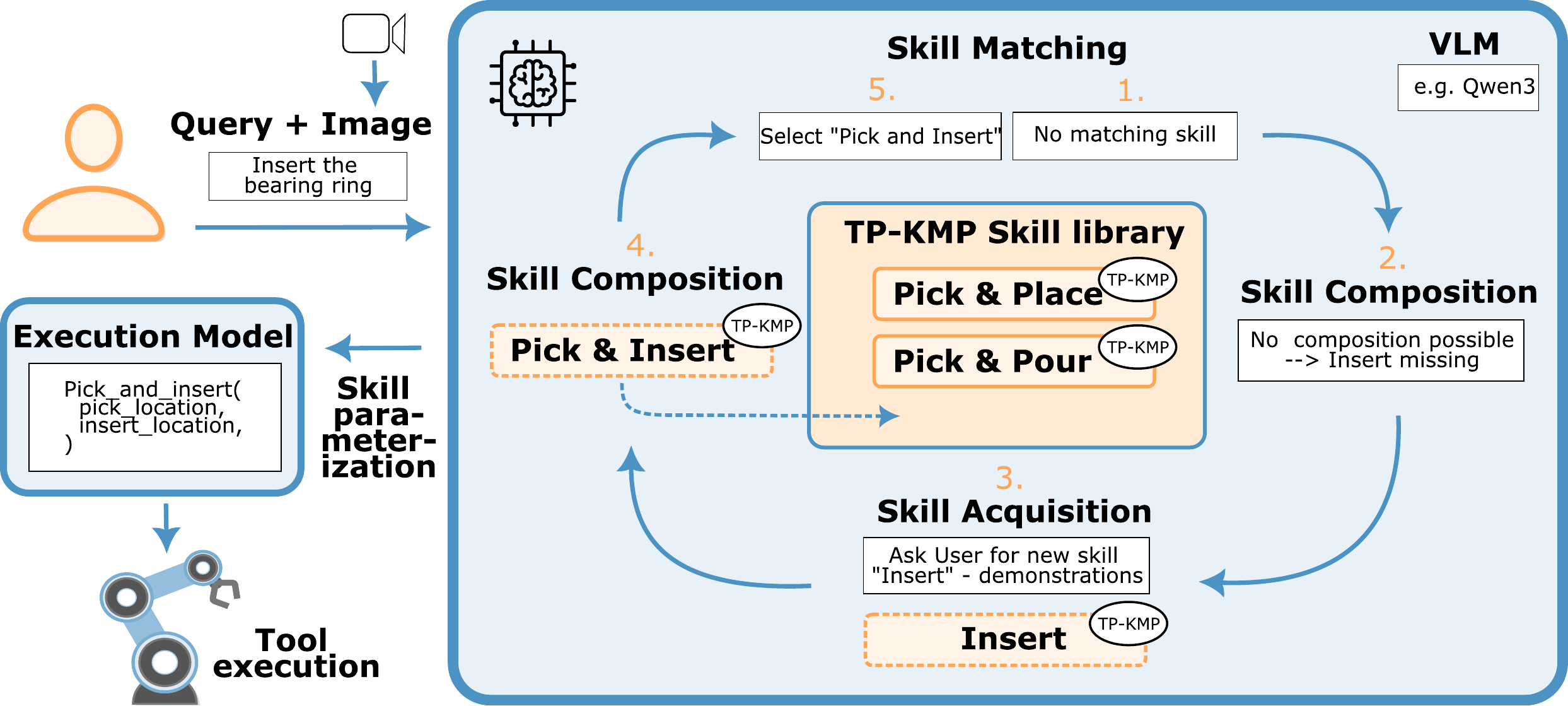}}\hfill
	\raisebox{-0.5\height}{\includegraphics[width=0.39\textwidth]{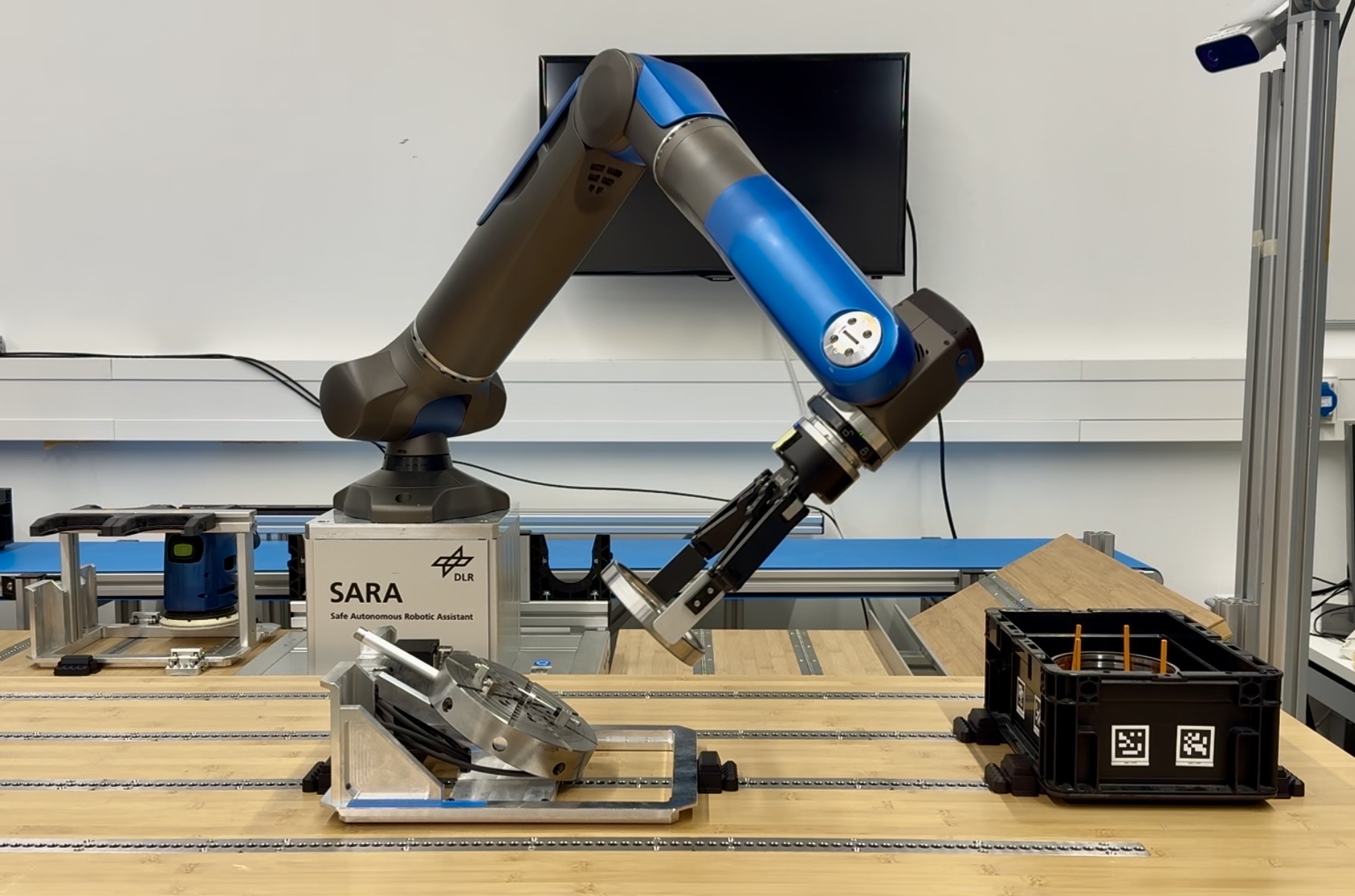}}
	\caption{\textbf{Left:} execution pipeline illustrated for the command ``Insert the bearing ring'': skill matching finds no match~(1), composition fails because an insert skill is missing~(2), the system acquires the missing skill via new demonstrations~(3), composition now succeeds and creates a fused Pick~\&~Insert skill~(4), which is then selected~(5). Selected skills are parameterized with detected object poses and executed via TP-KMP trajectory generation. \textbf{Right:} the robot performing a bearing ring insertion task.}
	\label{fig:execution_pipeline}
\end{figure}

Robot deployment in assembly environments requires programming methods that are both intuitive for non-expert users and data-efficient \cite{Schaal1999, Argall2009}.
Traditional robot programming---explicit waypoints or hardcoded control laws---is tedious, brittle to environmental variations, and scales poorly to the wide variety of tasks in manufacturing environments.
Recent vision-language-action (VLA) models offer end-to-end natural language interfaces but require tens of thousands to hundreds of thousands of demonstrations, extensive training times, and substantial compute \cite{Kim2025, Yuan2025, Neill2024}.
Conversely, task-parameterized imitation learning methods \cite{Calinon2016} achieve remarkable data efficiency: only 2--5 demonstrations per skill are required.
By learning skills locally in object coordinate frames (i.e., spatial reference frames defined by object poses---not image frames), they facilitate generalization to new object locations.
However, they lack natural language interfaces, requiring operators to programmatically specify which skills to execute based on the available objects, their locations, and task requirements.
This work bridges this gap by combining TP-KMPs \cite{Huang2019, Knauer2025} with pretrained vision-language models, achieving both data efficiency and natural language interaction.
The framework operates in three stages.
\textit{Learning stage:} Skills are acquired through kinesthetic teaching (2--5 demonstrations per skill), and the VLM automatically analyzes detected object types/positions to generate JSON schemas describing each skill's task parameters (object poses), preconditions, and semantic properties.
This automatic schema generation eliminates manual skill definition, enabling the VLM to reason about skill semantics without fine-tuning.
\textit{Execution stage (see \cref{fig:execution_pipeline}):} Users issue natural language commands, and the VLM selects appropriate skills by matching commands against skill schemas and binding detected objects to task parameters.
Crucially, the framework can synthesize novel continuous behaviors that were never individually demonstrated, by fusing local trajectory distributions from compatible skills through covariance-weighted composition.
Unlike symbolic composition, which chains discrete skills sequentially, this trajectory-level fusion generates new motion profiles where each skill's learned covariance determines when it dominates the fused trajectory, enabling smooth transitions without explicit sequencing logic.
A formal compatibility constraint (\cref{eq:compatibility_constraint}) guarantees that composition is only attempted when the skills' confidence regions do not conflict, preventing the unintended motion blending that arises from unconstrained superposition of movement primitives \cite{Saveriano2023}.
\textit{Capability expansion stage:} When neither existing skills nor valid compositions satisfy a user request, the system detects this capability gap and autonomously requests targeted demonstrations, enabling continuous skill library expansion while maintaining 2--5 demonstration efficiency.

This work makes the following key contributions:
\begin{enumerate}
	\item A modular architecture that bridges VLMs and TP-KMPs through automatically generated skill schemas, enabling data-efficient skill generation without VLM fine-tuning (\cref{sec:methodology}).
	\item A skill fusion mechanism that composes sub-skills from independently taught skills by leveraging TP-KMP covariance-weighted fusion, paired with a compatibility metric that determines when fusion yields coherent motion (\cref{sec:skill_composition}).
	\item An active skill acquisition loop that detects capability gaps via VLM reasoning, requests targeted demonstrations, and integrates newly learned skills into the skill library (\cref{sec:active_learning}).
\end{enumerate}
We validate our approach on a 7-DoF torque-controlled robot across four evaluation dimensions: object-type and pose generalization, skill composition, and active capability gap detection (\cref{sec:evaluation}).

\section{Related Work}
\label{sec:related_work}
\textbf{Task-Parameterized Imitation Learning (TP-IL):}
Task-parameterized approaches such as task-parameterized Gaussian mixture models (TP-GMMs) \cite{Calinon2014,Calinon2016} and TP-KMPs \cite{Huang2019,Knauer2025} encode robot skills relative to task-relevant reference frames (such as object poses), enabling generalization to new object configurations while requiring only 2--5 demonstrations per skill.
These methods provide inherent uncertainty quantification via covariance matrices, enabling probabilistic fusion to identify the most relevant reference frame at each moment of a task, and have been shown to support interactive learning through physical corrections \cite{Knauer2025}.

\textbf{Foundation Models in Robotics:}
Vision-language models offer remarkable zero-shot reasoning and visual grounding capabilities \cite{Radford2021,Vaswani2017}.
End-to-end VLAs \cite{Kim2025} integrate perception, reasoning, and action prediction but require extensive task-specific training data and sacrifice interpretability.
Layered approaches using separate perception and execution modules have been explored \cite{Driess2023}, but integration with traditional movement primitives remains limited.
Alternative approaches fine-tune vision-language foundation models for direct manipulation policy learning through multi-modal imitation learning \cite{Li2024}, demonstrating that VLM architectures can be adapted for action prediction when provided with hundreds of demonstrations, although requiring significant effort to deploy for specific tasks in industry.

\textbf{Skill Composition and Grounding:}
Existing language-grounded skill systems compose skills \textit{symbolically}: selecting and sequencing discrete primitives without modifying their trajectories \cite{Grannen2024, Tziafas2024, Gu2025}.
This limits them to behaviors already present as individual skills; they cannot generate novel continuous motions from existing ones.
Trajectory-level fusion via products of Gaussians is an established mechanism in movement primitives \cite{Paraschos2013, Calinon2016, Huang2019} and has been applied to uncertainty-aware skill blending \cite{Silverio2019} and library-based reproduction \cite{Oikonomou2022}.
However, these works neither integrate language-guided skill selection nor formalize when composition should be refused: unconstrained superposition of conflicting primitives produces averaged trajectories without practical utility \cite{Paraschos2013, Saveriano2023}.
Our approach combines VLM-driven skill selection with trajectory-level fusion under a formal compatibility constraint (\cref{eq:compatibility_constraint}) that guarantees coherent composed motion or rejects incompatible pairs.
Task-parameterized approaches have been extended for generalized skill learning \cite{Huang2018}, few-shot adaptation \cite{Zhu2022}, and incremental single-skill refinement \cite{Hoyos2016, Knauer2025}, but not for cross-skill composition.

\textbf{Gap in Literature:}
No existing framework jointly provides natural language grounding, trajectory-level skill composition with formal compatibility guarantees, and the data efficiency of task-parameterized learning.
Language-guided systems remain limited to symbolic sequencing of fixed primitives, while trajectory-level fusion methods lack language interfaces and criteria for rejecting incompatible compositions.
Our approach closes this gap; key differences are summarized in \cref{tab:related_work_comparison}, with an extended discussion in \cref{sec:appendix_gap_in_literature,sec:appendix_related_work}.


\section{Methodology}
\label{sec:methodology}

\subsection{Framework Overview}

The framework combines TP-KMPs with a pretrained VLM (Qwen3-VL-32B-Instruct) \cite{Qwen2025}, selected for local deployment without external API calls, important for industrial applications with data privacy requirements.
The system operates in two phases:
In the \textbf{learning phase} (\cref{sec:learning_phase}), kinesthetic demonstrations and perception-derived environment observations $\mathcal{E}$ are used to train TP-KMPs and build the skill library $\mathcal{S}$.
In the \textbf{execution phase} (\cref{sec:execution_phase}), natural language commands, current observations $\mathcal{E}$, and the skill library $\mathcal{S}$ are combined to select, parameterize, or compose skills.\footnote{The 2--5 demonstration count refers to motion skill learning; the perception pipeline requires a separate one-time training per object class using synthetic data.}
The system comprises three main functional components: (1) \textbf{Perception} for object pose estimation, (2) \textbf{VLM Reasoning} for skill selection, parameterization, and schema creation, and (3) \textbf{Trajectory Generation} using TP-KMPs.
The modular design enables flexible substitution of the VLM components (to \eg Pixtral, GPT) without modifying the TP-KMP backend.

\subsection{Execution Phase}\label{sec:execution_phase}
User commands (natural language) together with images from the workspace are processed by the VLM, which is operated according to the following decision tree:
\begin{enumerate}
	\item \textbf{Skill Matching}: Match the request to a skill in $\mathcal{S}$ (\cref{sec:learning_phase,sec:active_learning}) based on schema descriptions (e.g., ``pick up the apple'' matches a \texttt{grasp} skill if the schema specifies graspable objects); $P \leq N_{\text{obj}}$ detected objects are selected as reference frames.
	\item \textbf{Composition Feasibility}: If no single skill in $\mathcal{S}$ matches, check whether compatible skills can be composed (\cref{sec:skill_composition}). Composition feasibility is verified mathematically via the compatibility constraint (\cref{eq:compatibility_constraint}), not by the VLM.
	\item \textbf{Active Skill Acquisition}: If neither suffices, expand $\mathcal{S}$ via active learning (\cref{sec:active_learning}).
\end{enumerate}
\begin{wrapfigure}{r}{0.6\linewidth}
	\vspace{-13pt}
	\centering
	\includegraphics[width=\linewidth]{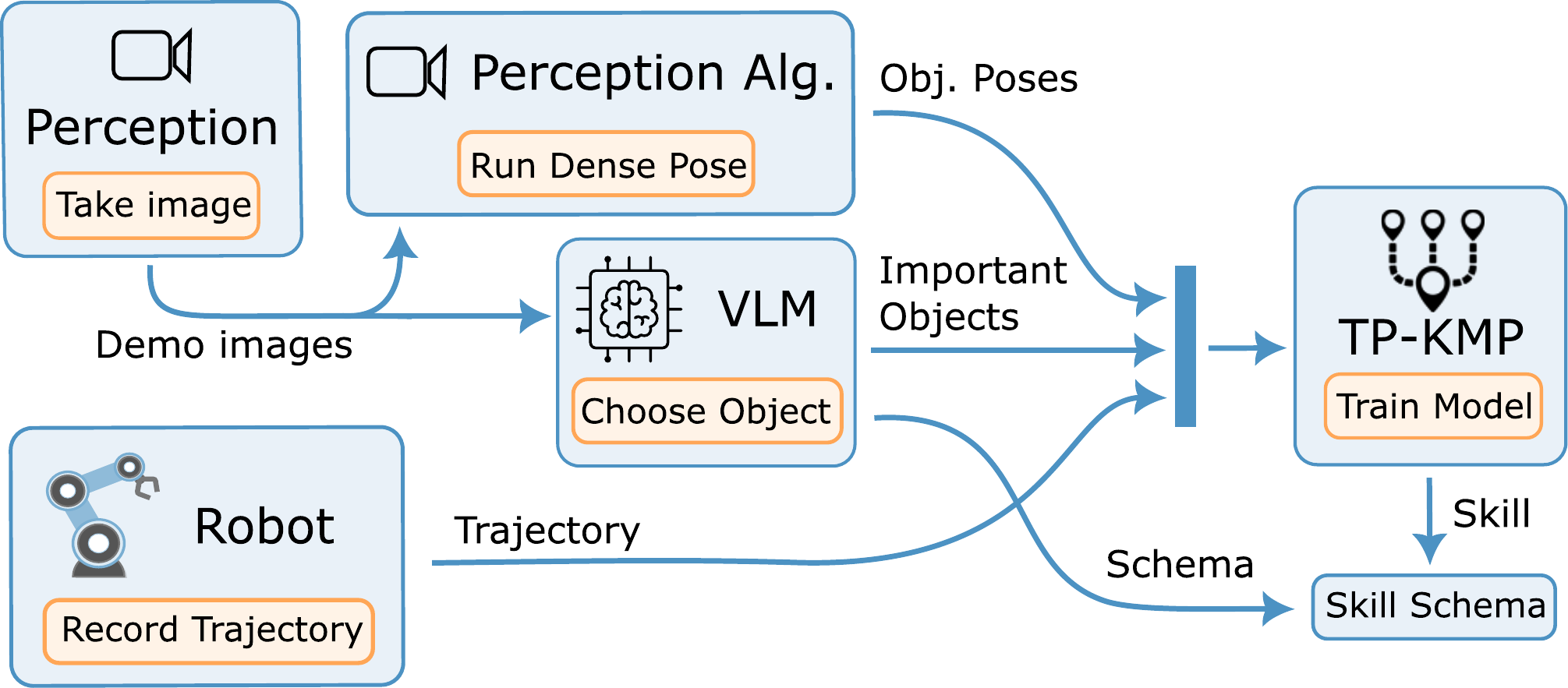}
	\caption{Skill schema creation in the \textbf{learning phase:} the VLM creates a schema as well as chooses relevant objects for the task from image input. The relevant objects, together with their pose estimation from the perception pipeline and the kinesthetic demonstration, are used to train a TP-KMP skill. Combined with its schema, the final skill is stored in the skill library.}
	\label{fig:training_pipeline}
	\vspace{-1em}
\end{wrapfigure}
All skill schemas are provided to the VLM simultaneously as tool definitions in a single prompt, alongside the user command and workspace image. The VLM's tool-calling mechanism \cite{Knauer2025, Wang2024, Qin2024} naturally handles skill selection by matching the request against available tool definitions.
Each skill is represented as a tool with attributes describing required objects and their interaction order, as demonstrated by the user.
The VLM's output is a tool call specifying which skill to execute and which detected objects to use as frames. Composed skills create new TP-KMPs by merging frames from existing skills (see \cref{sec:skill_composition}).
The complete execution phase workflow is illustrated in \cref{fig:execution_pipeline}.

\subsection{Learning Phase}\label{sec:learning_phase}

The full TP-KMP formalism is provided in \cref{sec:appendix_preliminaries}.
We define a \textit{skill} as a pair $(\tpkmp, \phi)$, where $\tpkmp = \{\localkmp\}_{\frameIndex=1}^{\amountOfFrames}$ is a trained TP-KMP with $\amountOfFrames$ local KMPs and $\phi$ is a VLM-generated schema encoding task parameters, interaction order, and semantic constraints.
The \textit{skill library} $\mathcal{S} = \{(\tpkmp_k, \phi_k)\}_{k=1}^{K}$ collects all learned skills available for execution and composition.

\textbf{Demonstration Collection.}
A user provides $\amountOfDatapoints = 2 \ldots 5$ kinesthetic demonstrations per skill, yielding a dataset $\mathcal{D} = \{\{\inputVariable_{\trajectoryIndex,\datapointIndex},\outputVariable_{\trajectoryIndex,\datapointIndex}\}_{\trajectoryIndex=1}^\trajectoryLength\}_{\datapointIndex=1}^\amountOfDatapoints$, where $\inputVariable \in \mathbb{R}^{\inputDimension}$ is the $\inputDimension$-dimensional phase variable, $\outputVariable \in \mathbb{R}^{\outputDimension}$ the $\outputDimension$-dimensional end-effector 6D pose with gripper state, and $\trajectoryLength$ the trajectory length, recorded at 100\,Hz.
Concurrently, the perception pipeline (\cref{sec:appendix_perception}) produces environment observations $\mathcal{E} = \{(\frameOrigin_{i}, \frameRotationMatrix_{i}), \bm{d}_{i}, \ell_{i}\}_{i=1}^{N_{\text{obj}}}$ for each detected object, comprising 6D pose (position $\frameOrigin_i$ and orientation $\frameRotationMatrix_i$), bounding box dimensions $\bm{d}_i = [w_i, h_i, d_i]^\top$ (width, height, depth), and semantic label~$\ell_i$.

\textbf{TP-KMP Training.}
From $\mathcal{E}$, $\amountOfFrames \leq N_{\text{obj}}$ task-relevant objects are selected as coordinate reference frames $\{(\localFrameOrigin, \localFrameRotationMatrix)\}_{\frameIndex=1}^{\amountOfFrames}$.
The demonstrations in $\mathcal{D}$ are projected into each frame's local coordinate system via~\cref{eq:local_projection}, and a local KMP $\localkmp = \{\kmpInput, \localkmpMean, \localkmpCovariance\}_{\kmpIndex=1}^{\amountOfKMP}$ encoding $\amountOfKMP$ local reference distributions is trained per frame, yielding the TP-KMP $\tpkmp = \{\localkmp\}_{\frameIndex=1}^{\amountOfFrames}$.
This frame-relative encoding enables learned movements to generalize to new object configurations by updating frame poses at execution time.

\textbf{Schema Generation.}
The VLM receives the workspace image and detected objects from $\mathcal{E}$, and generates a schema $\phi$ formatted as a JSON tool definition~\cite{Wang2024, Qin2024, Mai2026LLM}.
The schema specifies the number and type of required objects (\eg one graspable object and one surface) and a semantic skill label (\eg ``grasp'', ``pour'') assigned from visual context alone, without requiring user-provided task names.
Since the VLM does not observe the demonstrated trajectory, labels may reflect scene expectations rather than the actual motion; a self-verification pass mitigates such errors and improves schema quality (\cref{sec:evaluation}).
At execution time, the schema $\phi$ enables the VLM to select skills and bind specific object instances to the $\amountOfFrames$ reference frames (\cref{sec:execution_phase}), \eg ``\textit{Pick up the apple}'', ``\textit{Pour into the box}'', or ``\textit{Insert the ring into the station}''; an example schema is provided in \cref{lst:skill_schema}.
This eliminates the need for manual skill specification or fine-tuning.
The complete learning procedure is detailed in \cref{alg:skill_learning} and illustrated in \cref{fig:training_pipeline}.

\subsection{Skill Composition}
\label{sec:skill_composition}
We generate skills that explicitly encode object requirements, enabling the VLM to reason about which detected objects to use as reference frames through direct tool call arguments.
This is necessary because static tool calls, where VLMs infer skill behavior solely from skill names \cite{Ichter2023}, proved insufficient in our experiments. In scenes with multiple objects, the VLM could not specify which objects should serve as reference frames.
We propose a probabilistic mechanism for combining existing skills to create new behaviors.\footnote{The current evaluation validates pairwise composition. Since composed TP-KMPs are standard TP-KMPs, incremental chaining is possible in principle (see \cref{sec:discussion}).}
Given a skill library $\mathcal{S}$ containing an arbitrary number of TP-KMPs, the VLM first selects two candidate skills from the full library based on task semantics.
A new TP-KMP is then created by selecting one local KMP from each source skill and fusing them using the same product-of-Gaussians mechanism as in \cref{eq:tp_kmp_mean_and_covariance}.
This composition is only valid when the local KMPs satisfy the compatibility constraint (\cref{eq:compatibility_constraint}); incompatible skills (e.g., two local KMPs with overlapping high-confidence regions) cannot be meaningfully composed.

During execution, the $P$ selected objects are used as task parameters through their corresponding $\frameOrigin^{\local}, \frameRotationMatrix^{\local}$. The TP-KMP is evaluated in each object's local coordinate system and transformed to the global frame following \cref{eq:coordinate_transformation_mean_and_covariance}.
The selected TP-KMP generates a trajectory from the covariance-weighted fusion of frame-local predictions (\cref{eq:tp_kmp_mean_and_covariance}).
A Cartesian impedance controller \cite{Hogan1984} executes the trajectory in task space (see \cref{sec:appendix_impedance} for details).

\begin{wrapfigure}{r}{0.32\linewidth}
	\vspace{-2em}
	\centering
	\includegraphics[width=\linewidth]{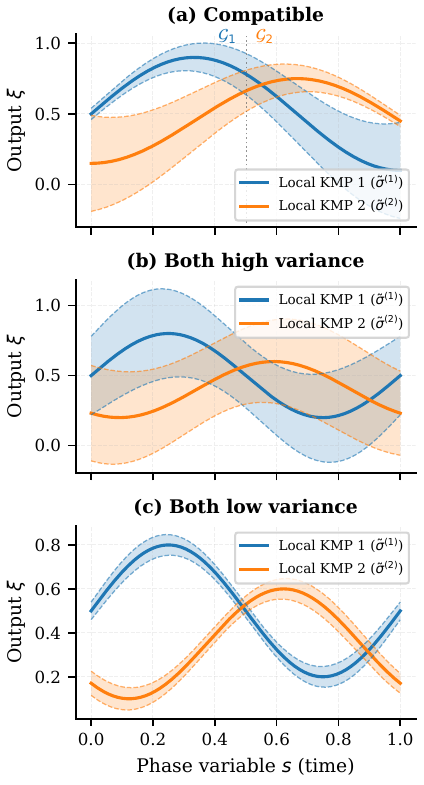}
	\caption{Compatibility constraint examples. (a)~Compatible: complementary variance creates non-overlapping dominant regions. (b)--(c)~Incompatible: both KMPs have uniformly high or low variance, preventing skill fusion.}
	\label{fig:compatibility_constraints}
	\vspace{-1em}
\end{wrapfigure}

\textbf{Compatibility Constraints.}
The compatibility constraint is a pre-check that validates whether two local KMPs can be composed. Here, $P = 2$ denotes the number of local KMPs being composed (one selected from each source skill), each associated with a distinct object coordinate frame, consistent with the general notation where $P$ is the number of task-relevant reference frames. For this pair to be compatible, the input domain (phase variable $\inputVariable$, here time $t$) must partition into $P$ non-overlapping regions $\mathcal{G}_1, \ldots, \mathcal{G}_P$, each dominated by exactly one local KMP.
For each region $\mathcal{G}_j$ and all $t \in \mathcal{G}_j$:
\begin{equation}
\label{eq:compatibility_constraint}
\forall q \in \{1,..,P\}\backslash\{\pi(j)\},\, \forall o \in \{1,..,\outputDimension\}:\;
\tilde{\sigma}_{o,t}^{(q)} > \tilde{\sigma}_{o,t}^{(\pi(j))} + \tau,
\end{equation}
where $\tilde{\sigma}_{o,t}^{(p)}$ is the standard deviation (square root of diagonal covariance element) for local KMP $p$ at time $t$ and output dimension $o$, and $\pi(j)$ is the local KMP with lowest uncertainty in region $j$.
When this constraint is satisfied (with margin $\tau > 0$, empirically $\tau = 0.01$), each local KMP dominates exactly one temporal region, and the fusion in \cref{eq:tp_kmp_mean_and_covariance} yields coherent motion with clean skill transitions. When violated, composition is rejected.
A detailed discussion of physical constraints affecting compatibility is provided in \cref{sec:appendix_compatibility}.
\Cref{fig:compatibility_constraints} illustrates compatibility examples.

\begin{figure*}[t]
\begin{minipage}[t]{0.45\textwidth}
\hrule height 0.8pt
\captionof{algorithm}{Skill Learning Procedure}
\label{alg:skill_learning}
\hrule height 0.4pt
\begin{algorithmic}[1]
\Require User confirmation, RGB-D sensor
\Ensure New skill added to $\mathcal{S}$
\State Robot enters gravity compensation mode
\State User demonstrates skill ($M = 2$--$5$ demonstrations)
\State Record $\mathcal{D} = \{(\inputVariable_{\datapointIndex}, \outputVariable_{\datapointIndex})\}_{\datapointIndex=1}^{\amountOfDatapoints}$ at 100 Hz with RGB-D observations $\mathcal{E}$
\State Detect objects $\to$ reference frames $(\bm{b}^{(p)}, \bm{A}^{(p)})$ for $p = 1, \ldots, P$
\State Train TP-KMP on $\mathcal{D}$ using frames from $\mathcal{E}$
\State VLM generates JSON schema
\State $\mathcal{S} \leftarrow \mathcal{S} \cup \{\text{new skill}\}$
\end{algorithmic}
\vspace{7.5pt}\hrule height 0.8pt
\end{minipage}%
\hfill\vrule width 0.4pt\hfill
\begin{minipage}[t]{0.54\textwidth}
\hrule height 0.8pt
\captionof{algorithm}{Skill Composition Procedure}
\label{alg:skill_composition}
\hrule height 0.4pt
\begin{algorithmic}[1]
\Require User request, TP-KMP skill library $\mathcal{S}$
\Ensure Composed TP-KMP or trigger skill learning
\State VLM selects candidate skills $A, B \in \mathcal{S}$ via schema matching
\State VLM selects frame indices $p_A \in \{1,\ldots,P_A\}$, $p_B \in \{1,\ldots,P_B\}$ based on task semantics
\State Verify $(\Theta^{p_A}, \Theta^{p_B})$ satisfy \cref{eq:compatibility_constraint}
\If{compatible}
    \State Create TP-KMP from $\Theta^{p_A}$ and $\Theta^{p_B}$ ($P = 2$)
    \State VLM generates composite skill + semantics
    \State \Return new TP-KMP
\Else
    \State \Return trigger \cref{alg:skill_learning}
\EndIf
\end{algorithmic}
\hrule height 0.8pt
\end{minipage}
\end{figure*}
The composition process is detailed in \cref{sec:appendix_composition_process} and summarized in \cref{alg:skill_composition}.

\subsection{Active Skill Acquisition}
\label{sec:active_learning}

When neither existing skills nor their compositions satisfy a user request, the framework enters an active learning workflow to acquire the missing skill.
This mechanism enables continuous, user-driven skill library expansion while maintaining data efficiency.
Our approach triggers a \textbf{Demo Request Generation}: the VLM generates a natural language request explaining why the skill cannot be executed and asking the user to demonstrate the motion.
Example: \textit{``I don't have a skill for inserting. Please demonstrate by physically guiding the end-effector.''}
Upon user confirmation, the skill is acquired via \cref{alg:skill_learning} and added to the library ($\mathcal{S} \leftarrow \mathcal{S} \cup \{(\tpkmp, \phi)\}$) including its auto-generated schema, making it immediately available for selection and composition.

\section{Evaluation}
\label{sec:evaluation}

\subsection{Experimental Setup}
The framework is tested on a 7-DoF collaborative robot \cite{Iskandar2020}
equipped with a robotiq gripper and static RGB-D camera positioned at the side of the workspace.
Impedance control (\cref{sec:appendix_impedance}, \cref{eq:impedance_control}) is configured with stiffness $\bm{K}_p = 750$ N/m (position) and $250$ Nm/rad (orientation), while damping $\bm{K}_d$ is calculated dynamically based on the robot configuration \cite{Iskandar2023}, selected to provide compliant interaction while maintaining trajectory tracking accuracy, consistent with \cite{Knauer2025}.
TP-KMP hyperparameters follow \cite{Knauer2025}.
Object detection and 6D pose estimation follow the perception pipeline described in \cref{sec:appendix_perception}.
All kinesthetic demonstrations were performed by a robotics expert.
The experiments are also shown in the accompanying video.

\begin{wraptable}{r}{0.68\linewidth}
	\centering
	\caption{Summary of evaluation results. Each skill was trained from 4 kinesthetic demonstrations.}
	\label{tab:evaluation_summary}
	\small
	\begin{tabular}{lcc}
		\toprule
		\textbf{Evaluation overview} & \textbf{Success Rate} & \textbf{Trials} \\
		\midrule
		\textbf{Object Generalization} (\cref{sec:object_gen_eval}) & 90.9\% & 40/44 \\
		\textbf{Pose Generalization} (vision, \cref{sec:pose_gen_eval}) & 79.3\% & 23/29 \\
		\textbf{Pose Generalization} (manual, \cref{sec:pose_gen_eval}) & 100\% & 29/29\\
		\textbf{Skill Composition} (\cref{sec:skill_comp_eval}) & 100\% & 16/16 \\
		\textbf{Active Skill Acq.} -- Gap Det. (\cref{sec:active_learn_eval}) & 95\% & 19/20 \\
		\textbf{Active Skill Acq.} -- Ring Insert. (\cref{sec:active_learn_eval}) & 73.3\% & 11/15 \\
		\bottomrule
	\end{tabular}
\end{wraptable}

\textbf{Evaluation Coverage:}
The experimental evaluation covers multiple dimensions of framework capability:
\textit{Object generalization} (\cref{sec:object_gen_eval}) tests whether pick-and-place skills demonstrated on one picked object transfer to different, unseen objects across 5 YCB objects (N=44 pairwise trials).
\textit{Pose generalization} (\cref{sec:pose_gen_eval}) tests the pick-and-pour task, using the same objects as during training, across 15 different spatial configurations.
\textit{Skill composition} (\cref{sec:skill_comp_eval}) validates the composition mechanism using pre-defined schemas.
\textit{Active skill acquisition} (\cref{sec:active_learn_eval}) tests capability gap detection and integration of newly acquired skills, including a precision bearing ring insertion task.
\textit{VLA baseline comparison} (\cref{sec:vla_baseline_eval}) provides a direct real-robot comparison against $\pi_{0.5}$ \cite{PhysicalIntelligence2025} fine-tuned on the same robot across the same object and pose generalization benchmarks, complemented by a qualitative comparison in terms of data efficiency and generalization with additional VLA approaches based on reported results (\cref{sub:qualitative_comparison}).
An overview of results from \cref{sec:object_gen_eval}--\cref{sec:active_learn_eval} is given in \cref{tab:evaluation_summary}. Detailed experimental protocols and additional evaluation figures are provided in \cref{sec:appendix_experimental_protocols,sec:appendix_eval_figures}.

\subsection{Object and Pose Generalization}
\label{sec:object_gen_eval}
\label{sec:pose_gen_eval}

\textbf{Object Generalization:} Each of the 5 YCB objects was used as a training object, and the learned skill was tested on all 5 objects, yielding N=44 pairwise trials. Overall success rate: 90.9\% (40/44 trials; see \cref{fig:object_gen}).
Successful generalization (40 trials): Skills successfully generalized to all tested object categories despite size, shape, and material differences.
Failures (4 trials): Attributed to two causes: size incompatibilities where learned grasping offsets became invalid (e.g., bearing ring with larger trained objects) and fragile object deformation (cracker box).
Additional details, including object combinations, used objects, and example trials, are provided in \cref{sec:appendix_experimental_protocols,sec:appendix_eval_figures}.

\textbf{Pose Generalization:} \textit{With automatic vision-based pose estimation:} 79.3\% success rate (23/29 trials), \textit{with manual pose specification:} 100\% success rate (all failures resolved with accurate poses).
Root cause of failures: Vision system limitations due to object occlusion.
This demonstrates that skill generalization to new poses is robust when pose estimates are accurate, but perception reliability is the primary bottleneck for fully autonomous execution. Note that pose generalization failures stem from perception errors, not size differences, unlike object generalization where size mismatches caused grasping failures.
Additional details, including example trials, are provided in \cref{sec:appendix_experimental_protocols,sec:appendix_eval_figures}.

\subsection{Skill Composition and Active Acquisition}
\label{sec:skill_comp_eval}
\label{sec:active_learn_eval}

\textbf{Composition with Existing Skills:}
Two base skills were composed: (1) grasp apple and place on plate, and (2) grasp potted meat can and pour into cracker box. The composition mechanism combines the grasp phase of one skill with the placement phase of the other, enabling novel pick-and-place combinations. We tested all 4 picking objects $\times$ 4 placing objects, yielding 16 trials. Composition Success Rate: 100\% (16/16 trials), demonstrating robust skill synthesis across all object combinations (see \cref{sec:appendix_experimental_protocols} for full protocol).

\textbf{Experimental Protocol:}
To demonstrate how active learning enables composition when the required skill is not yet in the library, we evaluated a bearing ring insertion scenario that requires sub-millimeter tolerances in a real manufacturing context, exceeding simple tabletop manipulation. The user requests a grasp-and-insert task, but the library lacks an \textit{insert} skill.
The pick-and-place skill from \cref{sec:skill_comp_eval} (trained on grasping an apple and placing on a plate) is reused to grasp a bearing ring.
However, the skill library initially lacks an \textit{insert} skill, requiring the framework to autonomously detect this capability gap and request demonstrations to acquire the missing skill.
The evaluation tests the complete active learning workflow: (1) capability gap detection when a user requests an unavailable composition, (2) successful integration of newly acquired skills into the composition mechanism, and (3) robustness of composed skills across spatial configurations.

\textbf{Results:}
\textbf{Phase 1 - Capability Gap Detection: 95\% (19/20 requests)}
When the user requests ``insert the bearing ring into the measurement station,'' the VLM analyzes available skills and detects that no insert skill exists in the library.
The framework correctly identified this capability gap in 95\% of trials (19/20 requests), with one false negative where the system attempted using a pick-and-place skill, and generated semantically appropriate demonstration requests (e.g., ``I don't have a skill for inserting. Please demonstrate by physically guiding the end-effector.'').
Upon user confirmation, the system collected demonstrations and trained the \textit{insert ring} skill, integrating it into the library.
Additional details, including covariance analysis and trajectory fusion, are provided in \cref{sec:appendix_covariance_analysis}.

\textbf{Phase 2 - Robustness of Newly Acquired and Composed Skills:} The grasp-and-insert skill was tested across 15 spatial configurations with varying measurement station positions and orientations (see \cref{sec:appendix_experimental_protocols}).
It achieved 73.3\% (11/15 trials), with all 4 failures at $180^\circ$ rotated configurations where the required insertion motion is fundamentally opposite to learned trajectories.

\subsection{Comparison with $\pi_{0.5}$ VLA}
\label{sec:vla_baseline_eval}

We fine-tuned $\pi_{0.5}$ \cite{PhysicalIntelligence2025} on the same robot and evaluated it on the same object generalization (\cref{sec:object_gen_eval}) and pose generalization (\cref{sec:pose_gen_eval}) benchmarks as our approach.
Multiple models were trained with different hyperparameters to ensure a fair evaluation.
As shown in \cref{tab:pi05_comparison}, with 5 demonstrations matching our demonstration budget, $\pi_{0.5}$ achieves 0\% autonomous success rate across all tasks: while the model shows clear task intention, it lacks the precision required for autonomous grasping.
With 50 demonstrations, $\pi_{0.5}$ reaches 86.4\% on object generalization, though it failed on the bearing ring: after fine-tuning on a fixed set of objects, $\pi_{0.5}$ did not attempt to grasp the visually dissimilar bearing ring despite the language instruction, instead moving toward objects more similar to its training set. It matches our approach at 79.3\% on pose generalization, where both approaches share the same camera occlusion bottleneck.

\begin{table}[tb]
\centering
\caption{Direct comparison on the same robot platform. Our approach uses 4 demonstrations per skill; $\pi_{0.5}$ was fine-tuned with 5 and 50 demonstrations respectively. Manual pose specification (---) is not applicable to end-to-end VLAs.}
\label{tab:pi05_comparison}
\small
\begin{tabular}{lccc}
\toprule
\textbf{Evaluation} & \textbf{Ours (4 demos)} & \textbf{$\pi_{0.5}$ (5 demos)} & \textbf{$\pi_{0.5}$ (50 demos)} \\
\midrule
Object Gen. (\cref{sec:object_gen_eval}) & 90.9\% & 0\% & 86.4\% \\
Pose Gen. (vision, \cref{sec:pose_gen_eval}) & 79.3\% & 0\% & 79.3\% \\
Pose Gen. (manual, \cref{sec:pose_gen_eval}) & 100\% & --- & --- \\
\bottomrule
\end{tabular}
\end{table}

\section{Discussion}
\label{sec:discussion}

Our framework demonstrates strong performance across all evaluation dimensions: 90.9\% on object generalization, 100\% on skill composition, and 95\% on capability gap detection for active acquisition of a precision industrial insertion skill.
Analysis of remaining failures reveals two distinct sources: perception errors (accounting for the gap between 79.3\% and 100\% in pose generalization) and orientation limits at $180^\circ$ rotations.
The direct comparison with $\pi_{0.5}$ (\cref{sec:vla_baseline_eval}) highlights a key trade-off: while $\pi_{0.5}$ matches our pose generalization performance at 79.3\%, it requires 50 demonstrations (compared to our 4) and 18 hours of fine-tuning on an H200 GPU, compared to seconds of TP-KMP training on a consumer CPU. With the same demonstration budget of 5, $\pi_{0.5}$ achieves 0\% autonomous success. This suggests our approach offers a viable alternative for specialized industrial applications where the object set is known, data collection is costly, and high precision is required.

\textbf{Limitations.}
Despite the promising results in \cref{sec:evaluation}, our approach presents some limitations.
Skill composition is currently limited to pairwise combinations satisfying the compatibility constraint (\cref{eq:compatibility_constraint}).
The perception pipeline assumes static object poses captured at the start of each skill, and depends on accurate 6D pose estimation; the current single-camera setup contributes to occlusion-related failures. For potential solutions, see \cref{sec:appendix_limitations}.
Finally, unlike end-to-end VLA approaches that can generalize to novel object categories zero-shot, our approach requires training the 6D pose estimator per object class: a 3D model is used to generate synthetic training data via BlenderProc \cite{Denninger2023}, and the pose estimator is trained on this data, a cost not captured by the 2--5 demonstration count for motion learning. However, in controlled production environments the object set is relatively stable while manipulation tasks change frequently; the one-time perception training cost is thus amortized over long-term deployment. Moreover, recent approaches enable 3D model reconstruction from only a single image \cite{Long2024}, reducing this overhead further.

\section{Conclusion}
\label{sec:conclusion}

This work bridges data-efficient, human-centered robot skill learning and natural language interfaces by combining TP-KMPs with pretrained VLMs through automatic skill schema generation, probabilistic skill composition with formal compatibility constraints, and active capability expansion.
Real-world validation confirms robust generalization across objects, poses, and heterogeneous skill compositions, with perception accuracy as the primary bottleneck (\cref{sec:evaluation}).
By achieving 2--5 demonstration efficiency with natural language grounding and principled composition criteria, this framework opens robot programming to non-experts without sacrificing precision.


\acknowledgments{This work was partially funded by the DLR project ``ASPIRO''; the European Union's Horizon Research and Innovation Program under Grant 101136067 (INVERSE); the Federal Ministry for Economic Affairs and Climate Protection with DARP funds based on a decision by the German Bundestag and by the European Union -- NextGenerationEU; and partially supported by the German Federal Ministry of Research, Technology and Space (BMFTR) under the Robotics Institute Germany (RIG).}





\bibliography{bibliography}

@inproceedings{Xu2024,
      title={P-RAG: Progressive Retrieval Augmented Generation For Planning on Embodied Everyday Task},
      author={Xu, Weiye and Wang, Min and Zhou, Wengang and Li, Houqiang},
      booktitle={ACM International Conference on Multimedia (MM)},
      year={2024},
      publisher={ACM},
      doi={10.1145/3664647.3680661},
}

@misc{Din2025,
      title={LLM-Guided Task and Motion Planning using Knowledge-based Reasoning},
      author={Din, Muhayy Ud and Rosell, Jan and Akram, Waseem and Zaplana, Isiah and Roa, Maximo A and Hussain, Irfan},
      year={2025},
      eprint={2412.07493},
      archivePrefix={arXiv},
      primaryClass={cs.RO},
      doi={10.48550/arXiv.2412.07493},
      note={\doi{10.48550/arXiv.2412.07493}},
}

@misc{Lei2025,
      title={CLEA: Closed-Loop Embodied Agent for Enhancing Task Execution in Dynamic Environments},
      author={Lei, Mingcong and Wang, Ge and Zhao, Yiming and Mai, Zhixin and Zhao, Qing and Guo, Yao and Li, Zhen and Cui, Shuguang and Han, Yatong and Ren, Jinke},
      year={2025},
      eprint={2503.00729},
      archivePrefix={arXiv},
      primaryClass={cs.RO},
      doi={10.48550/arXiv.2503.00729},
      note={\doi{10.48550/arXiv.2503.00729}},
}

@inproceedings{Ichter2023,
      author={Ichter, Brian and Brohan, Anthony and Chebotar, Yevgen and Finn, Chelsea and Hausman, Karol and Herzog, Alexander and Ho, Daniel and Ibarz, Julian and Irpan, Alex and Jang, Eric and Julian, Ryan and Kalashnikov, Dmitry and Levine, Sergey and Lu, Yao and Parada, Carolina and Rao, Kanishka and Sermanet, Pierre and Toshev, Alexander T and Vanhoucke, Vincent and Xia, Fei and Xiao, Ted and Xu, Peng and Yan, Mengyuan and Brown, Noah and Ahn, Michael and Cortes, Omar and Sievers, Nicolas and Tan, Clayton and Xu, Sichun and Reyes, Diego and Rettinghouse, Jarek and Quiambao, Jornell and Pastor, Peter and Luu, Linda and Lee, Kuang-Huei and Kuang, Yuheng and Jesmonth, Sally and Joshi, Nikhil J. and Jeffrey, Kyle and Ruano, Rosario Jauregui and Hsu, Jasmine and Gopalakrishnan, Keerthana and David, Byron and Zeng, Andy and Fu, Chuyuan Kelly},
      booktitle={Conference on Robot Learning (CoRL)},
      title={Do As I Can, Not As I Say: Grounding Language in Robotic Affordances},
      year={2023},
      volume={205},
      series={Proceedings of Machine Learning Research},
      publisher={PMLR},
      url={https://proceedings.mlr.press/v205/ichter23a.html},
}

@inproceedings{Certo2025,
      title={Large Language Model-Based Robot Task Planning from Voice Command Transcriptions},
      author={Certo, Afonso and Martins, Bruno and Azevedo, Carlos and Lima, Pedro U.},
      booktitle={IEEE/RSJ International Conference on Intelligent Robots and Systems (IROS)},
      year={2025},
      url={https://ieeexplore.ieee.org/document/11246378},
}

@article{Schaal1999,
      title={Is imitation learning the route to humanoid robots?},
      author={Schaal, Stefan},
      journal={Trends in Cognitive Sciences},
      volume={3},
      number={6},
      pages={233--242},
      year={1999},
      doi={10.1016/S1364-6613(99)01327-3},
}

@inproceedings{Calinon2014,
      author={Calinon, Sylvain and Bruno, Danilo and Caldwell, Darwin G.},
      booktitle={IEEE International Conference on Robotics and Automation (ICRA)},
      title={A task-parameterized probabilistic model with minimal intervention control},
      year={2014},
      pages={3339--3344},
      doi={10.1109/ICRA.2014.6907339},
}

@article{Huang2019,
      title={Kernelized Movement Primitives},
      author={Huang, Yanlong and Rozo, Leonel and Silv{\'e}rio, Jo{\~a}o and Caldwell, Darwin G.},
      journal={International Journal of Robotics Research (IJRR)},
      volume={38},
      number={7},
      pages={833--852},
      year={2019},
      doi={10.1177/0278364919846363},
}

@inproceedings{Huang2018,
      title={Generalized Task-Parameterized Skill Learning},
      author={Huang, Yanlong and Silv{\'e}rio, Jo{\~a}o and Rozo, Leonel and Caldwell, Darwin G.},
      booktitle={IEEE International Conference on Robotics and Automation (ICRA)},
      year={2018},
      doi={10.1109/ICRA.2018.8461079},
}

@article{Zhu2022,
      title={Learning Task-Parameterized Skills From Few Demonstrations},
      author={Zhu, Jihong and Gienger, Michael and Kober, Jens},
      journal={IEEE Robotics and Automation Letters (RA-L)},
      volume={7},
      number={2},
      pages={4063--4070},
      year={2022},
      doi={10.1109/LRA.2022.3150013},
}

@inproceedings{Paraschos2013,
      author={Paraschos, Alexandros and Daniel, Christian and Peters, Jan and Neumann, Gerhard},
      title={Probabilistic Movement Primitives},
      booktitle={Advances in Neural Information Processing Systems (NeurIPS)},
      year={2013},
      url={https://proceedings.neurips.cc/paper/2013/hash/e53a0a2978c28872a4505bdb51db06dc-Abstract.html},
}

@article{Saveriano2023,
      author={Saveriano, Matteo and Abu-Dakka, Fares J. and Kramberger, Alja{\v{z}} and Peternel, Luka},
      title={Dynamic Movement Primitives in Robotics: A Tutorial Survey},
      journal={International Journal of Robotics Research (IJRR)},
      volume={42},
      number={13},
      pages={1133--1184},
      year={2023},
      doi={10.1177/02783649231201196},
}

@article{Calinon2016,
      title={A tutorial on Task-Parameterized Movement Learning and Retrieval},
      author={Calinon, Sylvain},
      journal={Intelligent Service Robotics},
      volume={9},
      number={1},
      pages={1--29},
      year={2016},
      doi={10.1007/s11370-015-0187-9},
}

@inproceedings{Radford2021,
      title={Learning Transferable Visual Models From Natural Language Supervision},
      author={Radford, Alec and Kim, Jong Wook and Hallacy, Chris and Ramesh, Aditya and Goh, Gabriel and Agarwal, Sandhini and Sastry, Girish and Askell, Amanda and Mishkin, Pamela and Clark, Jack and Krueger, Gretchen and Sutskever, Ilya},
      booktitle={International Conference on Machine Learning (ICML)},
      year={2021},
      volume={139},
      pages={8748--8763},
      publisher={PMLR},
      url={https://proceedings.mlr.press/v139/radford21a.html},
}

@inproceedings{Driess2023,
      title={PaLM-E: An Embodied Multimodal Language Model},
      author={Driess, Danny and Xia, Fei and Sajjadi, Mehdi S. M. and Lynch, Corey and Chowdhery, Aakanksha and Ichter, Brian and Wahid, Ayzaan and Tompson, Jonathan and Vuong, Quan and Yu, Tianhe and Huang, Wenlong and Chebotar, Yevgen and Sermanet, Pierre and Duckworth, Daniel and Levine, Sergey and Vanhoucke, Vincent and Hausman, Karol and Toussaint, Marc and Greff, Klaus and Zeng, Andy and Mordatch, Igor and Florence, Pete},
      booktitle={International Conference on Machine Learning (ICML)},
      year={2023},
      publisher={PMLR},
      url={https://proceedings.mlr.press/v202/driess23a.html},
}

@misc{Qwen2025,
      title={Qwen3 Technical Report},
      author={Qwen Team},
      year={2025},
      eprint={2505.09388},
      archivePrefix={arXiv},
      primaryClass={cs.CL},
      doi={10.48550/arXiv.2505.09388},
      note={\doi{10.48550/arXiv.2505.09388}},
}

@inproceedings{Vaswani2017,
      title={Attention Is All You Need},
      author={Vaswani, Ashish and Shazeer, Noam and Parmar, Niki and Uszkoreit, Jakob and Jones, Llion and Gomez, Aidan N. and Kaiser, Lukasz and Polosukhin, Illia},
      booktitle={Advances in Neural Information Processing Systems (NeurIPS)},
      year={2017},
      volume={30},
      pages={6000--6010},
      url={https://proceedings.neurips.cc/paper/2017/hash/3f5ee243547dee91fbd053c1c4a845aa-Abstract.html},
}

@inproceedings{Yin2025,
      title={In-Context Learning Enables Robot Action Prediction in LLMs},
      author={Yin, Yida and Wang, Zekai and Sharma, Yuvan and Niu, Dantong and Darrell, Trevor and Herzig, Roei},
      booktitle={IEEE International Conference on Robotics and Automation (ICRA)},
      year={2025},
      doi={10.1109/ICRA55743.2025.11128807},
}

@inproceedings{Wang2024,
      title={What Are Tools Anyway? A Survey from the Language Model Perspective},
      author={Wang, Zhiruo and Cheng, Zhoujun and Zhu, Hao and Fried, Daniel and Neubig, Graham},
      booktitle={Conference on Language Modeling (COLM)},
      year={2024},
      url={https://openreview.net/pdf?id=Xh1B90iBSR},
}

@article{Qin2024,
      title={Tool Learning with Foundation Models},
      author={Qin, Yujia and Hu, Shengding and Lin, Yankai and Chen, Weize and Ding, Ning and Cui, Ganqu and Zeng, Zheni and Huang, Yufei and Xiao, Chaojun and Han, Chi and others},
      journal={ACM Computing Surveys (CSUR)},
      volume={57},
      pages={101:1--101:40},
      year={2024},
      doi={10.1145/3704435},
}

@inproceedings{Schick2023,
      title={Toolformer: Language Models Can Teach Themselves to Use Tools},
      author={Schick, Timo and Dwivedi-Yu, Jane and Dess\`{i}, Roberto and Raileanu, Roberta and Lomeli, Maria and Hambro, Eric and Zettlemoyer, Luke and Cancedda, Nicola and Scialom, Thomas},
      booktitle={Advances in Neural Information Processing Systems (NeurIPS)},
      year={2023},
      volume={36},
      url={https://proceedings.neurips.cc/paper_files/paper/2023/hash/d842425e4bf79ba039352da0f658a906-Abstract-Conference.html},
}

@inproceedings{Huang2023,
      title={Towards Reasoning in Large Language Models: A Survey},
      author={Huang, Jie and Chang, Kevin Chen-Chuan},
      booktitle={Findings of the Association for Computational Linguistics: ACL 2023},
      year={2023},
      pages={1049--1065},
      doi={10.18653/v1/2023.findings-acl.67},
}

@inproceedings{Grannen2024,
      title={Vocal Sandbox: Continual Learning and Adaptation for Situated Human-Robot Collaboration},
      author={Grannen, Jennifer and Karamcheti, Siddharth and Mirchandani, Suvir and Liang, Percy and Sadigh, Dorsa},
      booktitle={Conference on Robot Learning (CoRL)},
      year={2024},
      volume={270},
      series={Proceedings of Machine Learning Research},
      publisher={PMLR},
      url={https://proceedings.mlr.press/v270/grannen25a.html},
}

@inproceedings{Zitkovich2023,
      title={RT-2: Vision-Language-Action Models Transfer Web Knowledge to Robotic Control},
      author={Zitkovich, Brianna and Yu, Tianhe and Xu, Sichun and Xu, Peng and Xiao, Ted and Xia, Fei and Wu, Jialin and Wohlhart, Paul and Welker, Stefan and Wahid, Ayzaan and Vuong, Quan and Vanhoucke, Vincent and Tran, Huong and Soricut, Radu and Singh, Anikait and Singh, Jaspiar and Sermanet, Pierre and Sanketi, Pannag R. and Salazar, Grecia and Ryoo, Michael S. and others},
      booktitle={Conference on Robot Learning (CoRL)},
      volume={229},
      pages={2165--2183},
      year={2023},
      publisher={PMLR},
      url={https://proceedings.mlr.press/v229/zitkovich23a.html},
}

@misc{Barreiros2025,
      title={A Careful Examination of Large Behavior Models for Multitask Dexterous Manipulation},
      author={Barreiros, Jose and Beaulieu, Andrew and Bhat, Aditya and Cory, Rick and Cousineau, Eric and Dai, Hongkai and Fang, Ching-Hsin and Hashimoto, Kunimatsu and Irshad, Muhammad Zubair and Itkina, Masha and Kuppuswamy, Naveen and Lee, Kuan-Hui and Liu, Katherine and McConachie, Dale and McMahon, Ian and Nishimura, Haruki and Phillips-Grafflin, Calder and Richter, Charles and Shah, Paarth and Srinivasan, Krishnan and Wulfe, Blake and Xu, Chen and Zhang, Mengchao and others},
      year={2025},
      eprint={2507.05331},
      archivePrefix={arXiv},
      primaryClass={cs.RO},
      doi={10.48550/arXiv.2507.05331},
      note={\doi{10.48550/arXiv.2507.05331}},
}

@article{Argall2009,
      title={A survey of robot learning from demonstration},
      author={Argall, Brenna D. and Chernova, Sonia and Veloso, Manuela and Browning, Brett},
      journal={Robotics and Autonomous Systems},
      volume={57},
      number={5},
      pages={469--483},
      year={2009},
      doi={10.1016/j.robot.2008.10.024},
}

@inproceedings{Wang2023,
      title={YOLOv7: Trainable Bag-of-Freebies Sets New State-of-the-Art for Real-Time Object Detectors},
      author={Wang, Chien-Yao and Bochkovskiy, Alexey and Liao, Hong-Yuan Mark},
      booktitle={IEEE/CVF Conference on Computer Vision and Pattern Recognition (CVPR)},
      year={2023},
      pages={7464--7475},
      doi={10.1109/CVPR52729.2023.00721},
}

@article{Besl1992,
      author={Besl, Paul J. and McKay, Neil D.},
      title={A Method for Registration of 3-{D} Shapes},
      journal={IEEE Transactions on Pattern Analysis and Machine Intelligence (TPAMI)},
      volume={14},
      number={2},
      pages={239--256},
      year={1992},
      doi={10.1109/34.121791},
}

@inproceedings{Strobl2008,
      author={Strobl, Klaus H. and Hirzinger, Gerd},
      title={More Accurate Camera and Hand-Eye Calibrations with Unknown Grid Pattern Dimensions},
      booktitle={IEEE International Conference on Robotics and Automation (ICRA)},
      year={2008},
      pages={1398--1405},
      doi={10.1109/ROBOT.2008.4543398},
}

@inproceedings{Sundermeyer2018,
      author={Sundermeyer, Martin and Marton, Zoltan-Csaba and Durner, Maximilian and Brucker, Manuel and Triebel, Rudolph},
      title={Implicit 3D Orientation Learning for 6D Object Detection from RGB Images},
      booktitle={European Conference on Computer Vision (ECCV)},
      year={2018},
      doi={10.1007/978-3-030-01231-1_43},
}

@article{Hogan1984,
      title={Impedance control of industrial robots},
      author={Hogan, Neville},
      journal={Robotics and Computer-Integrated Manufacturing},
      volume={1},
      number={1},
      pages={97--113},
      year={1984},
      doi={10.1016/0736-5845(84)90084-X},
}

@inproceedings{Iskandar2020,
      author={Iskandar, Maged and Ott, Christian and Eiberger, Oliver and Keppler, Manuel and Albu-Sch{\"a}ffer, Alin and Dietrich, Alexander},
      booktitle={IEEE/RSJ International Conference on Intelligent Robots and Systems (IROS)},
      title={Joint-Level Control of the DLR Lightweight Robot SARA},
      year={2020},
      pages={8903--8910},
      doi={10.1109/IROS45743.2020.9340700},
}

@article{Iskandar2023,
      author={Iskandar, Maged and Ott, Christian and Albu-Sch{\"a}ffer, Alin and Siciliano, Bruno and Dietrich, Alexander},
      journal={IEEE Robotics and Automation Letters (RA-L)},
      title={Hybrid Force-Impedance Control for Fast End-Effector Motions},
      year={2023},
      volume={8},
      number={7},
      pages={3931--3938},
      doi={10.1109/LRA.2023.3270036},
}

@article{Calli2017,
      title={Yale-CMU-Berkeley dataset for robotic manipulation research},
      author={Calli, Berk and Singh, Arjun and Bruce, James and Walsman, Aaron and Konolige, Kurt and Srinivasa, Siddhartha and Abbeel, Pieter and Dollar, Aaron M},
      journal={International Journal of Robotics Research (IJRR)},
      volume={36},
      number={3},
      pages={261--268},
      year={2017},
      doi={10.1177/0278364917700714},
}

@article{Knauer2025,
      title={Interactive Incremental Learning of Generalizable Skills with Local Trajectory Modulation},
      author={Knauer, Markus and Albu-Sch{\"a}ffer, Alin and Stulp, Freek and Silv{\'e}rio, Jo{\~a}o},
      journal={IEEE Robotics and Automation Letters (RA-L)},
      volume={10},
      number={4},
      pages={3398--3405},
      year={2025},
      doi={10.1109/LRA.2025.3542209},
}

@inproceedings{Petruzzellis2025,
      author={Petruzzellis, Flavio and Cornelio, Cristina and Lio, Pietro},
      booktitle={International Conference on Machine Learning (ICML)},
      title={Hierarchical Planning for Complex Tasks with Knowledge Graph-RAG and Symbolic Verification},
      year={2025},
      volume={267},
      series={Proceedings of Machine Learning Research},
      publisher={PMLR},
      url={https://proceedings.mlr.press/v267/petruzzellis25a.html},
}

@inproceedings{Li2024,
      title={Vision-Language Foundation Models as Effective Robot Imitators},
      author={Li, Xinghang and Liu, Minghuan and Zhang, Hanbo and Yu, Cunjun and Xu, Jie and Wu, Hongtao and Cheang, Chilam and Jing, Ya and Zhang, Weinan and Liu, Huaping and Li, Hang and Kong, Tao},
      booktitle={International Conference on Learning Representations (ICLR)},
      year={2024},
      url={https://openreview.net/forum?id=lFYj0oibGR},
}

@misc{Kagaya2024,
      title={RAP: Retrieval-Augmented Planning with Contextual Memory for Multimodal LLM Agents},
      author={Kagaya, Tomoyuki and Yuan, Thong Jing and Lou, Yuxuan and Karlekar, Jayashree and Pranata, Sugiri and Kinose, Akira and Oguri, Koki and Wick, Felix and You, Yang},
      year={2024},
      eprint={2402.03610},
      archivePrefix={arXiv},
      primaryClass={cs.LG},
      doi={10.48550/arXiv.2402.03610},
      note={\doi{10.48550/arXiv.2402.03610}},
}

@inproceedings{Kim2025,
      author={Kim, Moo Jin and Pertsch, Karl and Karamcheti, Siddharth and Xiao, Ted and Balakrishna, Ashwin and Nair, Suraj and Rafailov, Rafael and Foster, Ethan P and Sanketi, Pannag R and Vuong, Quan and Kollar, Thomas and Burchfiel, Benjamin and Tedrake, Russ and Sadigh, Dorsa and Levine, Sergey and Liang, Percy and Finn, Chelsea},
      title={OpenVLA: An Open-Source Vision-Language-Action Model},
      booktitle={Conference on Robot Learning (CoRL)},
      year={2025},
      volume={270},
      series={Proceedings of Machine Learning Research},
      pages={2679--2713},
      publisher={PMLR},
      url={https://proceedings.mlr.press/v270/kim25c.html},
}

@inproceedings{Yuan2025,
      author={Yuan, Wentao and Duan, Jiafei and Blukis, Valts and Pumacay, Wilbert and Krishna, Ranjay and Murali, Adithyavairavan and Mousavian, Arsalan and Fox, Dieter},
      title={RoboPoint: A Vision-Language Model for Spatial Affordance Prediction in Robotics},
      booktitle={Conference on Robot Learning (CoRL)},
      year={2025},
      volume={270},
      series={Proceedings of Machine Learning Research},
      pages={4005--4020},
      publisher={PMLR},
      url={https://proceedings.mlr.press/v270/yuan25c.html},
}

@inproceedings{Neill2024,
      author={O'Neill, Abby and Rehman, Abdul and Maddukuri, Abhiram and Gupta, Abhishek and others},
      booktitle={IEEE International Conference on Robotics and Automation (ICRA)},
      title={Open {X-Embodiment}: Robotic Learning Datasets and {RT-X} Models},
      pages={6892--6903},
      year={2024},
      doi={10.1109/ICRA57147.2024.10611477},
}

@Article{Denninger2023,
  author    = {Maximilian Denninger and Dominik Winkelbauer and Martin Sundermeyer and Wout Boerdijk and Markus Knauer and Klaus H. Strobl and Matthias Humt and Rudolph Triebel},
  title     = {BlenderProc2: A Procedural Pipeline for Photorealistic Rendering},
  journal   = {Journal of Open Source Software (JOSS)},
  year      = {2023},
  volume    = {8},
  number    = {82},
  pages     = {4901},
  doi       = {10.21105/joss.04901},
  publisher = {The Open Journal},
}

@InProceedings{Kim2025N2,
  author    = {Kim, Moo Jin and Finn, Chelsea and Liang, Percy},
  title     = {Fine-Tuning Vision-Language-Action Models: Optimizing Speed and Success},
  booktitle = {Robotics: Science and Systems (RSS)},
  year      = {2025},
  doi       = {10.15607/RSS.2025.XXI.017},
}

@InProceedings{Tziafas2024,
  author    = {Tziafas, Georgios and Kasaei, Hamidreza},
  title     = {Lifelong Robot Library Learning: Bootstrapping Composable and Generalizable Skills for Embodied Control with Language Models},
  booktitle = {IEEE International Conference on Robotics and Automation (ICRA)},
  year      = {2024},
  pages     = {515--522},
  doi       = {10.1109/ICRA57147.2024.10611448},
}

@InProceedings{Gu2025,
  author    = {Gu, Weiwei and Kondepudi, Suresh and Gupta, Anmol and Huang, Lixiao and Gopalan, Nakul},
  title     = {Continual Robot Skill and Task Learning via Dialogue},
  booktitle = {IEEE International Conference on Robotics and Automation (ICRA) Workshop on Human-Centered Robot Learning},
  year      = {2025},
  url       = {https://openreview.net/forum?id=r7PpkXMoVk},
}

@InProceedings{Silverio2019,
  author    = {Silv{\'e}rio, Jo{\~a}o and Huang, Yanlong and Abu-Dakka, Fares J. and Rozo, Leonel and Caldwell, Darwin G.},
  title     = {Uncertainty-aware imitation learning using kernelized movement primitives},
  booktitle = {IEEE/RSJ International Conference on Intelligent Robots and Systems (IROS)},
  year      = {2019},
  pages     = {90-97},
  doi       = {10.1109/IROS40897.2019.8967996},
}

@InProceedings{Oikonomou2022,
  author    = {Oikonomou, Paris and Dometios, Athanasios and Khamassi, Mehdi and Tzafestas, Costas S.},
  title     = {Reproduction of Human Demonstrations with a Soft-Robotic Arm based on a Library of Learned Probabilistic Movement Primitives},
  booktitle = {2022 International Conference on Robotics and Automation (ICRA)},
  year      = {2022},
  pages     = {5212--5218},
  doi       = {10.1109/ICRA46639.2022.9811627},
}

@misc{PhysicalIntelligence2025,
      title={$\pi_{0.5}$: a Vision-Language-Action Model with Open-World Generalization},
      author={Physical Intelligence and Kevin Black and Noah Brown and James Darpinian and Karan Dhabalia and Danny Driess and Adnan Esmail and Michael Equi and Chelsea Finn and Niccolo Fusai and Manuel Y. Galliker and Dibya Ghosh and Lachy Groom and Karol Hausman and Brian Ichter and Szymon Jakubczak and Tim Jones and Liyiming Ke and Devin LeBlanc and Sergey Levine and Adrian Li-Bell and Mohith Mothukuri and Suraj Nair and Karl Pertsch and Allen Z. Ren and Lucy Xiaoyang Shi and Laura Smith and Jost Tobias Springenberg and Kyle Stachowicz and James Tanner and Quan Vuong and Homer Walke and Anna Walling and Haohuan Wang and Lili Yu and Ury Zhilinsky},
      year={2025},
      eprint={2504.16054},
      archivePrefix={arXiv},
      primaryClass={cs.LG},
      doi={10.48550/arXiv.2504.16054},
      note={\doi{10.48550/arXiv.2504.16054}},
}

@article{Hoyos2016,
  author  = {Hoyos, Jose and Prieto, Flavio and Aleny{\`a}, Guillem and Torras, Carme},
  title   = {Incremental learning of skills in a task-parameterized gaussian mixture model},
  journal = {Journal of Intelligent \& Robotic Systems},
  volume  = {82},
  pages   = {81--99},
  year    = {2016},
  doi     = {10.1007/s10846-015-0290-3},
}

@INPROCEEDINGS{Long2024,
  author    = {Long, Xiaoxiao and Guo, Yuan-Chen and Lin, Cheng and Liu, Yuan and Dou, Zhiyang and Liu, Lingjie and Ma, Yuexin and Zhang, Song-Hai and Habermann, Marc and Theobalt, Christian and Wang, Wenping},
  booktitle = {2024 IEEE/CVF Conference on Computer Vision and Pattern Recognition (CVPR)},
  title     = {{Wonder3D: Single Image to 3D Using Cross-Domain Diffusion}},
  year      = {2024},
  pages     = {9970--9980},
  doi       = {10.1109/CVPR52733.2024.00951},
}

@inproceedings{Mai2026LLM,
  author    = {Mai, Tai and Sakagami, Ryo and Quere, Gabriel and Mesesan, George and Schuller, Robert and Fr{\"u}nd, Konrad and Vogel, J{\"o}rn and Hagengruber, Annette and Lee, Jinoh and D{\"o}mel, Andreas and Stulp, Freek and Bustamante, Samuel},
  title     = {{LLM} Tool Workflows for Robot Explainability and Natural Language Commanding},
  booktitle = {ICRA 2026 Workshop on Semantics for Reliable Robot Autonomy: From Environment Understanding and Reasoning to Safe Interaction},
  year      = {2026},
  url       = {https://openreview.net/forum?id=NUu9P1LwbT},
}

\clearpage

\appendix

\section{Supplementary Material}

This appendix provides supplementary material. \Cref{sec:appendix_gap_in_literature} provides the extended gap-in-literature discussion. \Cref{sec:appendix_related_work} provides additional related work discussion. \Cref{sub:qualitative_comparison} offers a qualitative comparison with VLA approaches. \Cref{sec:appendix_preliminaries} provides the full TP-KMP mathematical formalism. \Cref{sec:appendix_perception} describes the perception pipeline. \Cref{sec:appendix_impedance} details the impedance controller. \Cref{sec:appendix_compatibility} discusses compatibility constraints in detail. \Cref{sec:appendix_composition_process} describes the composition process and algorithm. \Cref{sec:appendix_experimental_protocols} provides the experimental protocols for all evaluation scenarios. \Cref{sec:appendix_eval_figures} collects additional evaluation figures. \Cref{sec:appendix_covariance_analysis} provides the covariance structure analysis for active learning composition. \Cref{sec:appendix_limitations} provides the full limitations discussion.

\subsection{Gap in Literature -- Extended Discussion}\label{sec:appendix_gap_in_literature}

While TP-IL methods provide data efficiency and geometric interpretability through task frames, they lack natural language interfaces for accessible user interaction.
Language-guided approaches \cite{Grannen2024,Tziafas2024,Gu2025} compose skills symbolically but do not operate at the trajectory level.
Conversely, trajectory-level fusion methods \cite{Silverio2019,Oikonomou2022} lack language grounding, automatic skill selection, and formal criteria for refusing incompatible compositions.
This gap motivates our approach: combining VLM-based multimodal grounding with probabilistic task-parameterized movement primitives enables a system that (1) adapts quickly to novel environments with minimal demonstrations, (2) reasons about uncertainty in both perception and execution, (3) maintains geometric interpretability through frame-relative representations, and (4) provides natural language interfaces.
Such integration addresses the core challenge of accessible, data-efficient robot programming for specialized industrial applications where demonstrations are expensive and task-specific adaptation is essential.

\subsection{Additional Related Work}\label{sec:appendix_related_work}

\textbf{Natural Language in Robot Control:}
Recent work combines foundation models with imitation learning via in-context learning (ICL) \cite{Yin2025}, embedding demonstrations directly in prompts.
End-to-end systems demonstrate the feasibility of direct voice-to-plan translation using lightweight fine-tuned Large-Language Models (LLMs) \cite{Certo2025}, achieving robustness to automatic speech recognition (ASR) errors through diverse beam search and error-augmented training.
However, these approaches are constrained by context window limitations and cannot accumulate skills for long-term library development.

\textbf{Language-Guided Continual Skill Learning:}
Several recent works combine language models with skill primitives for continual learning.
Vocal Sandbox \cite{Grannen2024} uses DMPs with LLM-based skill selection and also addresses capability gap detection.
Lifelong Robot Library Learning \cite{Tziafas2024} bootstraps composable skills via LLM-generated code.
Continual Skill and Task Learning via Dialogue \cite{Gu2025} acquires and composes learned visuo-motor policies through dialogue-based interaction.
These approaches compose skills \textit{symbolically}, selecting and sequencing discrete primitives rather than operating at the continuous trajectory level.
Trajectory-level fusion via products of Gaussians is an established mechanism in probabilistic and kernelized movement primitives \cite{Paraschos2013, Calinon2016, Huang2019}, and has been applied to uncertainty-aware skill blending \cite{Silverio2019} and library-based reproduction \cite{Oikonomou2022}.
However, these works do not integrate language-guided skill selection or formalize when composition should be refused.
Our work combines VLM-driven skill selection with a formal compatibility constraint that determines when trajectory-level fusion yields coherent motion, embedded in an active learning loop for skill library expansion.

\textbf{VLM Tool Calling and Reasoning:}
Tool calling enables language models to invoke external functions and APIs, significantly extending their capability beyond text generation \cite{Qin2024,Schick2023}.
Systematic surveys of tool use in foundation models \cite{Wang2024} highlight VLM success in tool selection and parameterization, particularly when tools are explicitly specified in schemas.
Grounding language models in robotic affordances represents a key challenge, with SayCan \cite{Ichter2023} pioneering the combination of LLM semantic knowledge with learned value functions to enable robots to select feasible skills for long-horizon tasks.
However, challenges persist in multi-step reasoning for complex task decomposition \cite{Huang2023}, especially when the reasoning must consider abstract compatibility constraints between tools.

Retrieval-augmented approaches \cite{Xu2024,Kagaya2024} achieve data efficiency via memory augmentation. Hierarchical planners \cite{Petruzzellis2025,Din2025} ensure correctness through symbolic verification. Closed-loop frameworks like CLEA \cite{Lei2025} address dynamic replanning through critic-based monitoring. However, these methods operate at the discrete action level without probabilistic uncertainty modeling over continuous movement primitives.

\begin{table}[tb]
	\centering
	\caption{Comparison of Language-Grounded Robot Skill Learning Approaches. Sequential composition chains discrete skills; probabilistic composition fuses continuous trajectory distributions. CLEA has partial active learning through critic-triggered replanning.}
	\label{tab:related_work_comparison}
		\small
		\begin{tabular}{lccccc}
			\toprule
			\textbf{Method} & \makecell{\textbf{Skill}\\\textbf{Execution}} & \makecell{\textbf{Sequential}\\\textbf{Composition}} & \makecell{\textbf{Probabilistic}\\\textbf{Composition}} & \makecell{\textbf{Active}\\\textbf{Learning}} \\
			\midrule
			SayCan \cite{Ichter2023} & \checkmark & $\times$ & $\times$ & $\times$ \\
			RAP \cite{Kagaya2024} & \checkmark & $\times$ & $\times$ & $\times$ \\
			CLEA \cite{Lei2025} & \checkmark & $\times$ & $\times$ & $\sim$ \\
			Vocal Sandbox \cite{Grannen2024} & \checkmark & \checkmark & $\times$ & \checkmark \\
			Lifelong Robot Lib. \cite{Tziafas2024} & \checkmark & \checkmark & $\times$ & \checkmark \\
			Cont. Skill via Dial. \cite{Gu2025} & \checkmark & $\sim$ & $\times$ & \checkmark \\
			\midrule
			\textbf{Our Approach} & \checkmark & \checkmark & \checkmark & \checkmark \\
			\bottomrule
		\end{tabular}
\end{table}

\begin{table}[htbp]
	\centering
	\caption{Positioning of our approach with respect to VLA models on manipulation tasks. In-domain and out-of-domain success rates for pick, place, and pour tasks. VLA pretraining data requirements are shown; fine-tuning requires fewer samples. $^\dagger$No pretraining; 2--5 task-specific demonstrations per skill. Discussed in the qualitative evaluation (\cref{sub:qualitative_comparison}).}
	\label{tab:vla_comparison}
	\small
		\begin{tabular}{lccc}
			\toprule
			\textbf{Method} & \makecell{\textbf{In-Domain}\\\textbf{Success}} & \makecell{\textbf{Out-of-Domain}\\\textbf{Success}} & \makecell{\textbf{Pretraining}\\\textbf{Data}} \\
			\midrule
			RT-2 \cite{Zitkovich2023} & - & 30--55\% & $>$100k demos \\
			OpenVLA \cite{Kim2025} & - & 60--87\% & $>$100k demos \\
			LBM \cite{Barreiros2025} & 50--95\% & 40--80\% & $>$10k demos \\
			OpenVLA-OFT+\cite{Kim2025N2} & 51--100\% & 51--100\% & 20--300 demos \\
			\midrule
			\textbf{Ours (auto. perception)} & \textbf{100\%} & 79--91\% & \textbf{none}$^\dagger$ \\
			\bottomrule
		\end{tabular}
\end{table}

\subsection{Qualitative Comparison to VLAs}\label{sub:qualitative_comparison}

We provide \textit{qualitative comparative statements} between our framework and recent vision-language-action (VLA) models on manipulation tasks (see also \cref{tab:vla_comparison}).

\textbf{Generalization Performance:}
On in-domain tasks (pick, place, pour), VLA models achieve 50--95\% success rates \cite{Barreiros2025}, while our approach achieves 100\% due to the TP-KMP backend's deterministic trajectory generation.
For out-of-domain generalization, we evaluated object generalization (40/44 trials, 90.9\%) and pose generalization (23/29 trials, 79.3\%), both using automatic perception.
Comparative VLA results: \cite{Barreiros2025} reports 40--80\%, OpenVLA \cite{Kim2025} 60--87\%, and RT-2 \cite{Zitkovich2023} 30--55\%.
Note that the nature of generalization differs fundamentally: VLA out-of-domain evaluation typically involves novel object categories and semantic task variations, whereas our evaluation considers geometric generalization, \ie novel poses and positions of object classes for which the perception pipeline has been trained.
Our modular pipeline succeeds on the bearing ring because the 6D pose estimator was trained on it, while $\pi_{0.5}$'s end-to-end fine-tuning overrides its pretrained language grounding, reducing generalization to objects seen during fine-tuning.

\textbf{Data Efficiency:}
VLA models require large-scale pretraining datasets (tens of thousands to hundreds of thousands of demonstrations) \cite{Kim2025,Zitkovich2023}, though fine-tuning requires considerably fewer samples.
Our framework requires only 2--5 demonstrations per skill without pretraining of the motion generation pipeline. However, unlike VLAs which can generalize to novel objects zero-shot, deploying on new object classes requires training the perception module (see limitations in \cref{sec:discussion}).



\subsection{TP-KMP Formalism}\label{sec:appendix_preliminaries}

Similar to \cite{Knauer2025}, we denote a set of $\amountOfDatapoints$ demonstrations as $\mathcal{D} = \{\{\inputVariable_{\trajectoryIndex,\datapointIndex},\outputVariable_{\trajectoryIndex,\datapointIndex}\}_{\trajectoryIndex=1}^\trajectoryLength\}_{\datapointIndex=1}^\amountOfDatapoints$, where $\inputVariable\in\mathbb{R}^{\inputDimension}$ is the input variable (typically time), $\outputVariable \in \mathbb{R}^{\outputDimension}$ is the output variable (e.g., end-effector position), and $\trajectoryLength$ is the trajectory length.

\textbf{Kernelized Movement Primitives (KMPs)} \cite{Huang2019} provide a probabilistic approach for learning robot skills from limited human demonstrations.
From the demonstrations $\mathcal{D}$, a KMP encodes a skill from a reference trajectory distribution $\kmp = \left\lbrace \kmpInput, \kmpMean, \kmpCovariance \right\rbrace^\amountOfKMP_{\kmpIndex=1}$ comprising $\amountOfKMP$ Gaussians with means $\kmpMean$ and covariances $\kmpCovariance$, computed using Gaussian Mixture Regression.
Unlike point-wise trajectory learning, this approach naturally captures demonstration uncertainty: regions with high variance across demonstrations produce large covariances, while consistent demonstrations result in tight covariance estimates.
Starting from this reference trajectory distribution, a KMP generates a Gaussian distribution $\mathcal{N}(\mean, \covariance)$ for any query input, interpolating between the learned reference points.
These uncertainty measures are essential for skill composition, as they identify which trajectory regions are most confident and serve as anchors for merging multiple skills without retraining.

\textbf{Task-parameterized kernelized movement primitives (TP-KMPs)} \cite{Calinon2016, Knauer2025} extend KMPs by encoding trajectories relative to task-relevant object reference frames.
Each frame $\frameIndex = 1, \ldots, \amountOfFrames$ (typically representing objects in the workspace) is described by \textit{task parameters}: a position $\frameOrigin^{\local} \in \mathbb{R}^3$ and orientation $\frameRotationMatrix^{\local} \in SO(3)$ (a rotation matrix) representing the object's coordinate system with respect to a common reference frame (e.g., the robot base).
During learning, demonstrations recorded in the global reference frame are projected into each object's local coordinate system:
\begin{equation}
\label{eq:local_projection}
\outputVariable^{\local} = (\frameRotationMatrix^{\local})^{-1}\left(\outputVariable - \frameOrigin^{\local}\right),
\end{equation}
where $\outputVariable^{\local}$ represents the end-effector trajectory relative to frame $\frameIndex$.
For each frame, the local datasets are modeled probabilistically, yielding a Gaussian distribution $\mathcal{N}(\localMean, \localCovariance)$ for every input $\inputVariable$.
During execution, each KMP provides local predictions in the respective frame's coordinate system (from the local data learned during the learning phase).
The current task parameters at execution time may differ from those observed during learning due to object movement; we denote them as $\frameRotationMatrixAtTimestep^{\local}, \frameOriginAtTimestep^{\local}$.
These local distributions are transformed to the global coordinate frame using these current task parameters:
\begin{equation}
\label{eq:coordinate_transformation_mean_and_covariance}
\hat{\mean}^{\local}_{\timeStep} = \frameRotationMatrixAtTimestep^{\local} \localMean_{\timeStep} + \frameOriginAtTimestep^{\local};
\hat{\covariance}^{\local}_{\timeStep} = \frameRotationMatrixAtTimestep^{\local} \localCovariance_{\timeStep} (\frameRotationMatrixAtTimestep^{\local})^{\top}.
\end{equation}

Then, distributions from different coordinate systems are fused using a product of Gaussians, resulting in a global prediction.
For a query at time $\timeStep$, the fused distribution $\mathcal{N}(\mean_{\timeStep}, \covariance_{\timeStep})$ in the global frame is obtained by:

\begin{equation}
\label{eq:tp_kmp_mean_and_covariance}
\mean_{\timeStep} = \covariance_{\timeStep} \sum_{\frameIndex=1}^{\amountOfFrames} (\hat{\covariance}^{(\frameIndex)}_{\timeStep})^{-1} \hat{\mean}^{(\frameIndex)}_{\timeStep};
\covariance_{\timeStep} = \left(\sum_{\frameIndex=1}^{\amountOfFrames} (\hat{\covariance}^{(\frameIndex)}_{\timeStep})^{-1}\right)^{-1},
\end{equation}
where $\hat{\mean}^{(\frameIndex)}_{\timeStep}$ and $\hat{\covariance}^{(\frameIndex)}_{\timeStep}$ are the mean and covariance predictions from frame $\frameIndex$, transformed to the global coordinate frame according to \cref{eq:coordinate_transformation_mean_and_covariance}.
This fusion mechanism naturally weights frame contributions by inverse covariance: frames with low uncertainty (small covariance) contribute more strongly to the global prediction, while uncertain frames have minimal influence.
This weighting scheme automatically identifies the most confident frame at each point in the trajectory, enabling spatial partitioning of frame dominance without explicit region assignment, and favors models with consistent demonstrations that generalize to new situations.
The approach allows learned skills to generalize to new object configurations by updating frame poses at execution time.
The covariance matrices naturally weight the contributions of each frame, enabling compatibility assessment between skills, which forms the foundation of our skill composition mechanism.

\textbf{Perception Pipeline:}
We use YOLOv7 \cite{Wang2023} for object detection and an adversarial autoencoder \cite{Sundermeyer2018} for 6D pose estimation, both trained on the YCB dataset \cite{Calli2017}. Details are provided in \cref{sec:appendix_perception}.

\subsection{Perception Pipeline}\label{sec:appendix_perception}

Object detection employs YOLOv7 \cite{Wang2023} trained on the YCB dataset \cite{Calli2017} with confidence threshold $\kappa \geq 0.7$.
6D object pose estimation uses an adversarial autoencoder (AAE) architecture \cite{Sundermeyer2018} trained on synthetic data, generated with BlenderProc \cite{Denninger2023}, with domain randomization: lighting variations (intensity and color temperature), texture randomization, geometric transformations (scaling $\pm 5\%$, rotation $\pm 20\%$), and simulated occlusion through random box placement.
Hand-eye calibration is performed using standard checkerboard patterns \cite{Strobl2008}, followed by Iterative Closest Point (ICP) \cite{Besl1992} refinement for improved pose accuracy.

\begin{table}[htbp]
	\centering
	\caption{YCB objects~\cite{Calli2017} used in the object generalization evaluation (\cref{sec:object_gen_eval}). Objects span a range of sizes and geometries to test grasp transfer across different shapes.}
	\label{tab:ycb_object_table}
		\small
		\begin{tabular}{lc}
			\toprule
			\textbf{Object} & \textbf{YCB code} \\
			\midrule
			master chef can & YCB 002 \\
			cracker box & YCB 003 \\
			potted meat can & YCB 010 \\
			apple & YCB 013 \\
			plate & YCB 029 \\
			\bottomrule
		\end{tabular}
\end{table}

\begin{lstlisting}[
	language={},
	caption={Example skill schema automatically generated by the VLM during the learning phase for a pick-and-place skill. The schema defines task parameters, pre-/postconditions, primitive skill phases with their associated reference frames, and example usage context. This JSON tool definition enables the VLM to select and parameterize skills at execution time.},
	label=lst:skill_schema,
	basicstyle=\tiny\ttfamily\color{black},
	backgroundcolor=\color{pybackground},
	numberstyle=\tiny\color{pygray},
	breaklines=true,
	breakatwhitespace=true,
	columns=flexible,
	keepspaces=true,
	showstringspaces=false,
	tabsize=2,
	frame=single,
	frameround=tttt,
	framesep=5pt,
	rulecolor=\color{pygray!50},
	captionpos=b,
	numbers=left,
	stepnumber=1,
	numbersep=8pt,
	xleftmargin=17pt,
	xrightmargin=5pt,
	literate=
		*{:}{{{\color{black}:}}}{1}
		{,}{{{\color{black},}}}{1}
		{\{}{{{\color{pypurple}{\char`\{}}}}{1}
		{\}}{{{\color{pypurple}{\char`\}}}}}{1}
		{[}{{{\color{pypurple}[}}}{1}
		{]}{{{\color{pypurple}]}}}{1}
		{"skill\_name"}{{{\color{pyblue}"skill\_name"}}}{12}
		{"explanation"}{{{\color{pyblue}"explanation"}}}{13}
		{"example\_usage"}{{{\color{pyblue}"example\_usage"}}}{15}
		{"parameters"}{{{\color{pyblue}"parameters"}}}{12}
		{"object\_to\_pick"}{{{\color{pyblue}"object\_to\_pick"}}}{16}
		{"object\_to\_place"}{{{\color{pyblue}"object\_to\_place"}}}{17}
		{"type"}{{{\color{pyblue}"type"}}}{6}
		{"description"}{{{\color{pyblue}"description"}}}{13}
		{"object\_order"}{{{\color{pyblue}"object\_order"}}}{14}
		{"preconditions"}{{{\color{pyblue}"preconditions"}}}{15}
		{"postconditions"}{{{\color{pyblue}"postconditions"}}}{16}
		{"composition"}{{{\color{pyblue}"composition"}}}{13}
		{"primitive\_skills"}{{{\color{pyblue}"primitive\_skills"}}}{18}
		{"combinable\_with"}{{{\color{pyblue}"combinable\_with"}}}{17}
		{"PickAndPlace"}{{{\color{pyorange}"PickAndPlace"}}}{14}
		{"string"}{{{\color{pyorange}"string"}}}{8}
	]
{
  "skill_name": "PickAndPlace",
  "explanation": "This skill picks up one object and places
    it on top of another. The robot grasps object_to_pick,
    moves above object_to_place, and opens the gripper.",
  "example_usage": "Demonstrated picking an apple and placing
    it on a plate. Object sizes: Apple (0.075, 0.075, 0.075),
    Plate (0.258, 0.258, 0.01). Similar sizes recommended.",
  "preconditions": [
    "object_to_pick must be graspable and within reach",
    "object_to_place must have a flat surface for placement",
    "gripper must be empty before execution"
  ],
  "postconditions": [
    "object_to_pick is placed on object_to_place",
    "gripper is empty and open"
  ],
  "primitive_skills": [
      {"name": "Pick", "phase": "first", "frame": "object_to_pick",
       "description": "Grasping phase with high confidence
         near the object, low confidence during approach."},
      {"name": "Place", "phase": "second", "frame": "object_to_place",
       "description": "Placement phase with high confidence
         near target, low confidence during transport."}
    ],
  "parameters": {
    "object_to_pick": {
      "type": "string",
      "description": "The object to be picked up. Choose
        from the list of detected objects in the scene."
    },
    "object_to_place": {
      "type": "string",
      "description": "The target object serving as placement
        surface. Choose from detected objects in the scene."
    }
  },
  "object_order": ["object_to_pick", "object_to_place"]
}
\end{lstlisting}

\begin{figure}[tb]
	\centering
	\includegraphics[width=0.6\linewidth]{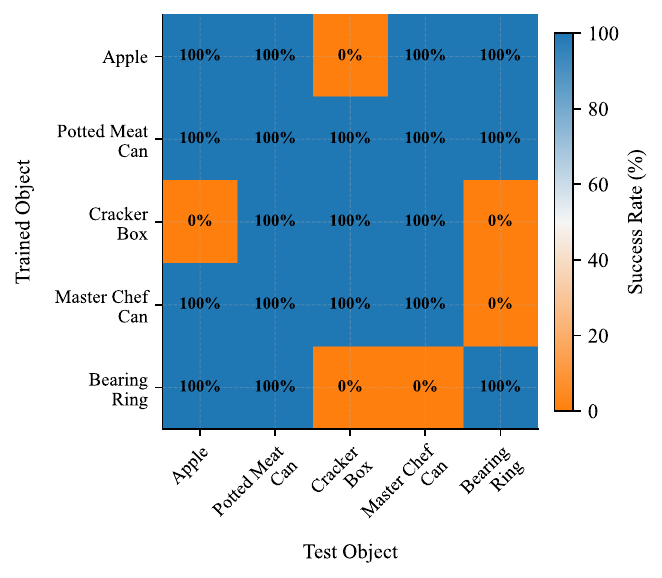}
	\caption{Object generalization success rates across object pairs.
		Each cell shows the success rate when the skill is trained on the row object and tested on the column object (N=44 total trials). Diagonal elements (training object matched at test time) approach 100\%, while performance degrades for smaller test objects. This reveals size incompatibility as the primary generalization limitation: skills trained on small objects transfer to larger ones, but not vice versa, as the learned grasp aperture may be too wide.}
	\label{fig:object_gen}
\end{figure}

\subsection{Impedance Control}\label{sec:appendix_impedance}

The Cartesian impedance controller computes desired task-space forces by modeling the end-effector as a mass-spring-damper system:
\begin{equation}
	\label{eq:impedance_control}
	\force = \bm{K}_p(\outputVariable_{\text{desired}} - \outputVariable_{\text{actual}}) + \bm{K}_d(\dot{\outputVariable}_{\text{desired}} - \dot{\outputVariable}_{\text{actual}}),
\end{equation}
where $\bm{K}_p$ and $\bm{K}_d$ are stiffness and damping matrices, and $\force$ is the desired task-space force command.
These task-space forces are mapped to joint torques internally by the robot's built-in impedance controller via the transposed Jacobian.
Torque limits and collision detection from joint sensors ensure safe human-robot collaboration throughout execution \cite{Iskandar2020}.

\subsection{Compatibility Constraints --- Detailed Discussion}\label{sec:appendix_compatibility}

Physical constraints of kinesthetic teaching, such as object stability limiting safe orientation ranges, result in learned covariance structures that may not naturally satisfy compatibility for all skill pairs.
For example, insertion tasks are naturally demonstrated with high confidence (low covariance) because tight tolerances are required, while grasping tasks can exhibit position variation but limited orientation variation.
This results in high-confidence regions that may overlap between skills, violating the compatibility constraint.
Composition is therefore not always possible without retraining; skills with incompatible covariance structures cannot be composed, and attempting to do so results in unintended motion blending, a fundamental limitation of movement primitive superposition, where contradictory motion directions average out to trajectories without practical utility as discussed in \cite{Paraschos2013, Saveriano2023}.

\subsection{Composition Process}\label{sec:appendix_composition_process}

Consider skill $A$ with $P_A$ local KMPs (one per frame, using notation from \cref{sec:appendix_preliminaries}: each frame $p$ has task parameters $(\bm{b}^{(p)}, \bm{A}^{(p)})$ and a learned distribution $\mathcal{N}(\bm{\mu}^{(p)}, \bm{\Sigma}^{(p)})$). Similarly, skill $B$ has $P_B$ local KMPs. For composition, the VLM selects one local KMP from each skill based on task semantics (e.g., selecting the ``grasp'' phase from skill $A$ and the ``place'' phase from skill $B$), identified by frame indices $p_A$ and $p_B$ respectively. The selected pair $(\Theta^{p_A}, \Theta^{p_B})$ yields a new TP-KMP with $P = 2$ local KMPs. The compatibility constraint (checked beforehand) ensures these two local KMPs have non-overlapping dominance regions, so the fusion produces coherent motion rather than unintended blending.
During trajectory generation, predictions from both selected local KMPs are fused via \cref{eq:tp_kmp_mean_and_covariance}.
The local KMP $\Theta^{p_A}$ dominates in temporal regions where it has low covariance, and similarly for $\Theta^{p_B}$, enabling automatic temporal partitioning and smooth skill transitions without retraining.

\subsection{Experimental Protocols}\label{sec:appendix_experimental_protocols}

\begin{figure}[!tb]
	\centering
	\includegraphics[width=\linewidth]{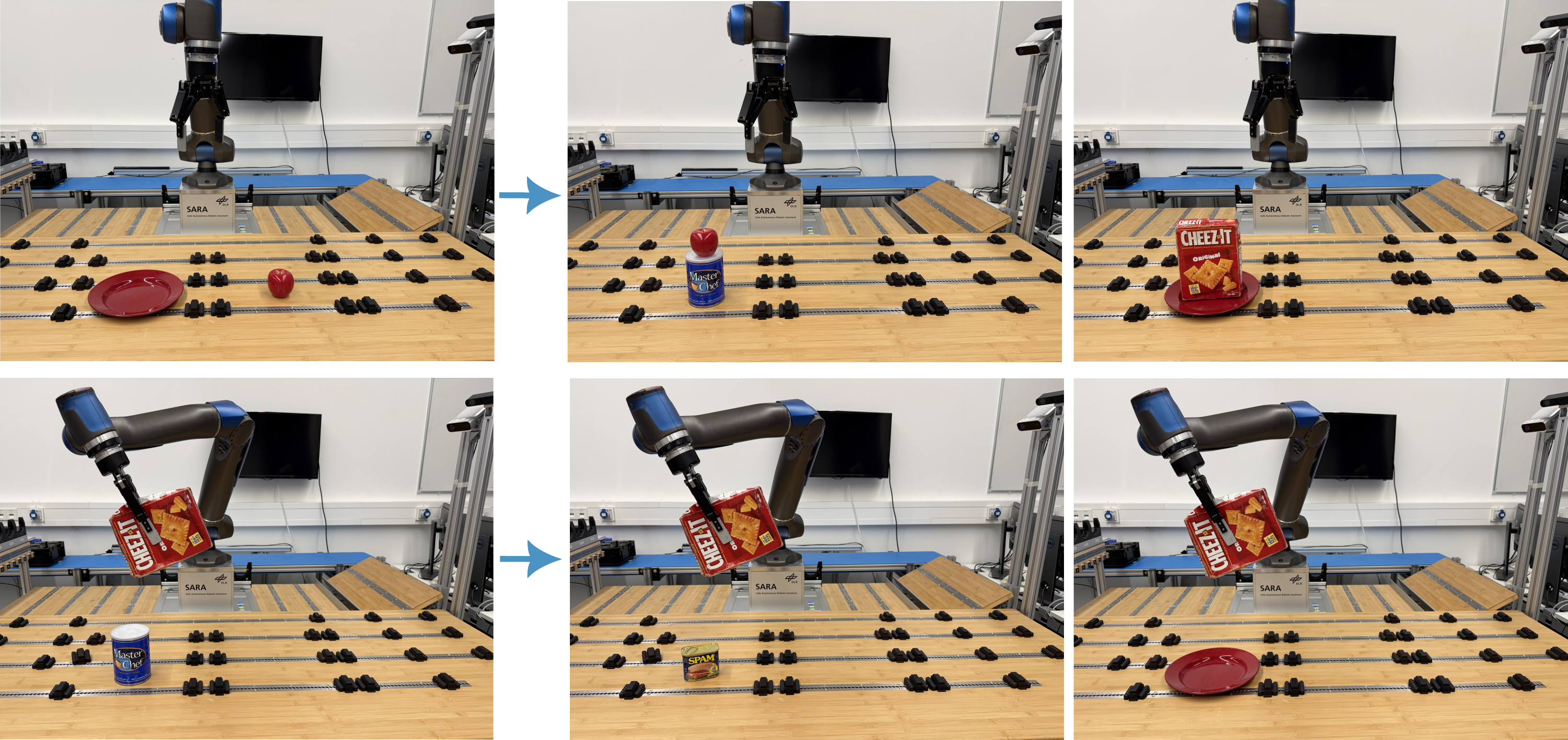}
	\caption{Examples of the object generalization evaluation: switching to unlearned objects for pick-place and pick-pour skills}
	\label{fig:eval_object_generalization}
\end{figure}

\textbf{Object Generalization:}
To evaluate generalization to novel object categories, we tested three skills across multiple object pairs.
The evaluation systematically tested generalization by training on one object and evaluating on different target objects.
Objects used for evaluation: YCB~013, 029, 002, 010, 003, and a bearing ring (see \cref{fig:eval_object_generalization}). For a YCB overview see \cref{tab:ycb_object_table}.
Tasks tested included:
\begin{itemize}
	\item \textbf{Pick skill:} Trained on (YCB 013), (010), (003), (002), and bearing ring individually; evaluated on generalization to other untrained objects.
	\item \textbf{Place skill:} Trained on (YCB 029) as target location; evaluated generalization to placing on top of (YCB 010), (003), (002) instead.
	\item \textbf{Pour skill:} Trained on (YCB 029) as target location; evaluated on pouring onto (YCB 010), (003), (002) instead.
\end{itemize}
For each task, multiple object-pair combinations were tested.
Success was defined as successful completion of the skill without collision or object deformation.

\textbf{Skill Composition:}
To validate the framework's ability to compose existing compatible skills, we tested heterogeneous skill combinations to demonstrate how skills can be combined to create novel behaviors without retraining.
Two base TP-KMPs are trained: (1) a grasp-and-place skill trained on grasping an apple (YCB 013) and placing it on a plate (YCB 029), and (2) a grasp-and-pour skill trained on picking a potted meat can (YCB 010) and pouring it into a cracker box (YCB 003).
The composition mechanism allows novel combinations: for instance, picking the potted meat can (from skill 2) and placing it on the plate (from skill 1).
The evaluation uses a systematic test matrix: 4 picking objects (YCB 013, 010, 003, 002) $\times$ 4 placing objects (YCB 029, 010, 003, 002), resulting in 16 different composition tests.

\begin{figure}[!tb]
	\centering
	\includegraphics[width=\linewidth]{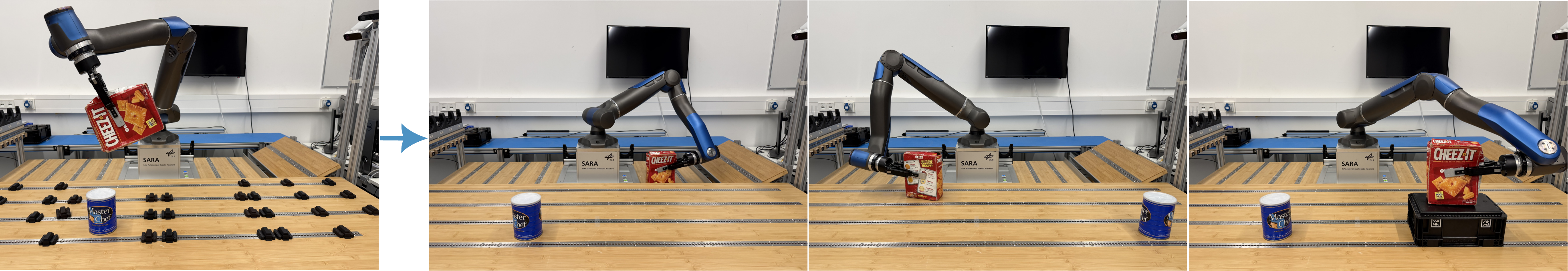}
	\caption{Examples for the pose generalization evaluation: trying different pick- and pour positions.}
	\label{fig:eval_pose_generalization}
\end{figure}

\begin{figure}[tb]
\centering
\annotategraphicsmulti[draw=none, fill=none]{%
  \includegraphics[width=0.65\linewidth]{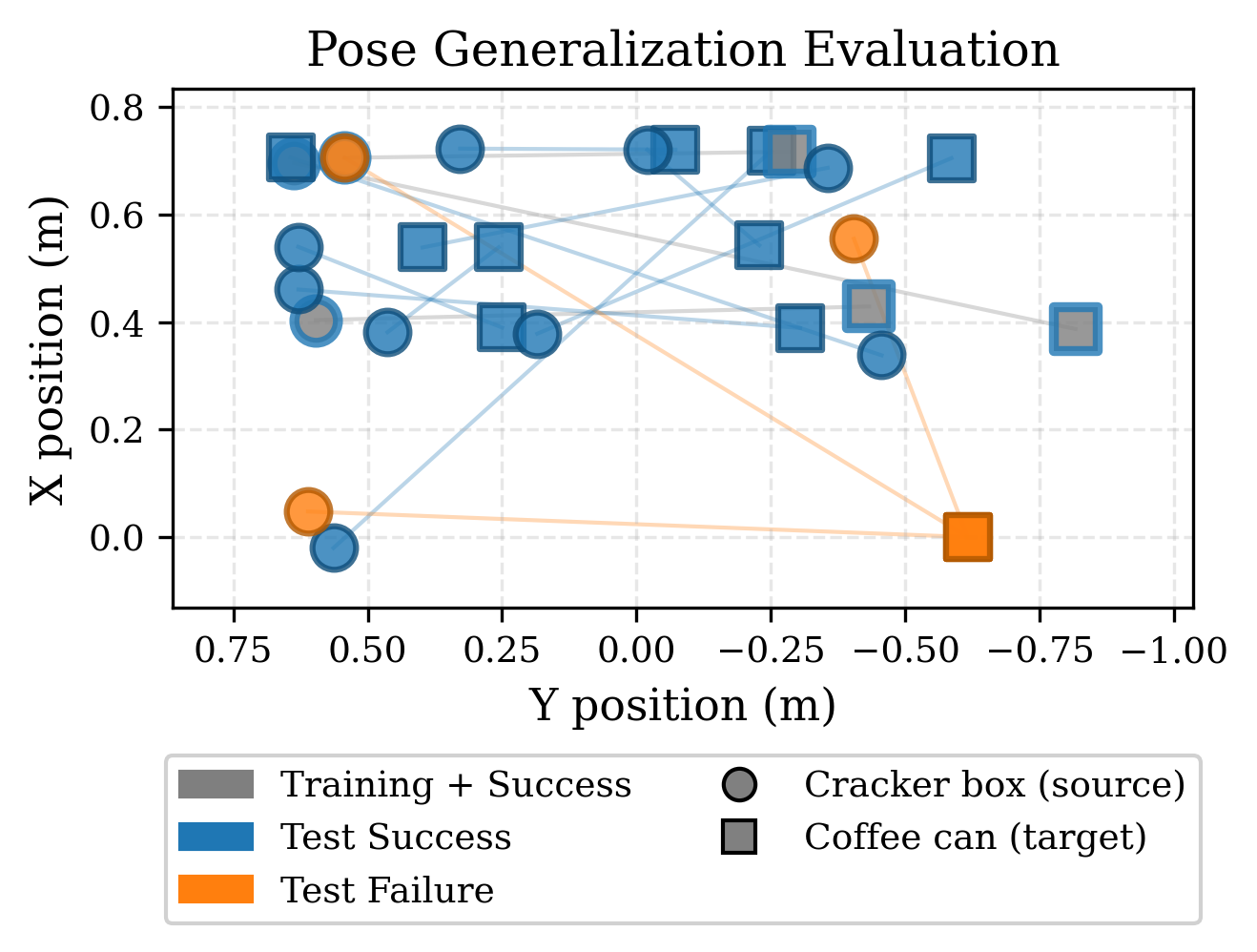}%
}{%
  \node[draw=none, fill=none] at (0.06, 0.34) {%
    \begin{tikzpicture}
      \node[rotate=40, scale=0.1] at (0,0) {\includegraphics{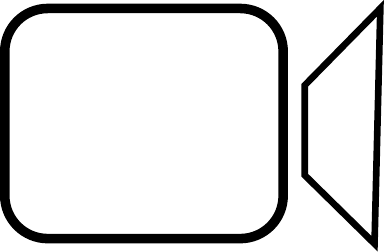}};
  \end{tikzpicture}%
  };
}
\caption{Spatial distribution of evaluated object positions for pose generalization. Blue markers indicate successful trials, orange markers indicate failures. Gray markers indicate training positions. Circles represent cracker box (source), squares represent master chef can (target).}
\label{fig:pose_gen_positions}
\end{figure}

\textbf{Pose Generalization:}
To evaluate robustness to variations in object pose and workspace configuration, we tested the grasp-and-pour skill (grasping the cracker box and pouring into a master chef can) across 15 different spatial configurations (see \cref{fig:eval_pose_generalization,fig:pose_gen_positions}).
Each configuration consisted of different 6D pose estimates for the source and target objects.
For each position configuration, the full grasp-and-pour motion was executed once.
Success was defined as: (1) successful gripper closure around cracker box, (2) collision-free transport to pouring position over master chef can, and (3) successful pouring motion.

\begin{figure}[tb]
\centering
\annotategraphicsmulti[draw=none, fill=none]{%
  \includegraphics[width=0.65\linewidth]{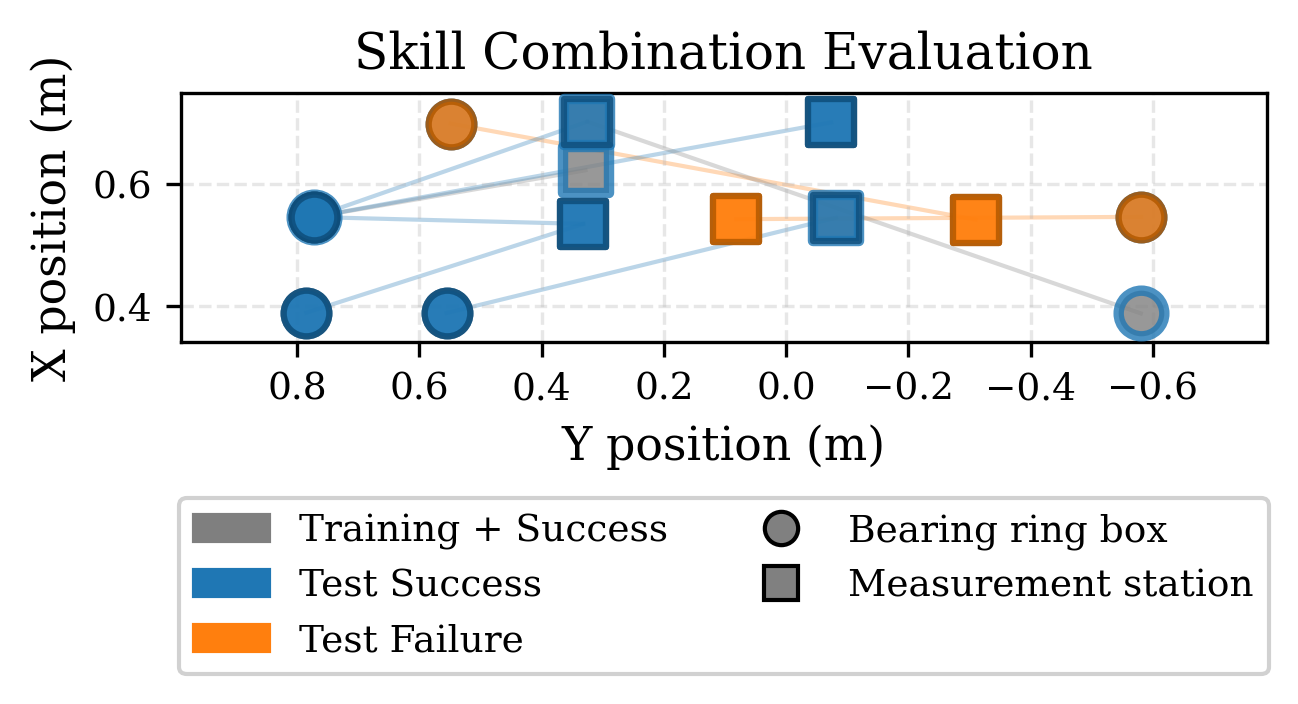}%
}{%
  \node[draw=none, fill=none] at (0.06, 0.34) {%
    \begin{tikzpicture}
      \node[rotate=30, scale=0.1] at (0,0) {\includegraphics{camera-svg}};
  \end{tikzpicture}%
  };
}
\caption{Spatial distribution of evaluated object positions for skill combination. Blue markers indicate successful trials, orange markers indicate failures. Gray markers indicate training positions. Circles represent bearing ring box, squares represent measurement station.}
\label{fig:skill_comb_positions}
\end{figure}

\textbf{Active Learning Phase 2 --- Robustness Evaluation:}
Having acquired the \textit{insert ring} skill through active learning, we now validate the robustness of the grasp-and-insert composed skill across varying spatial configurations.
The newly acquired and composed grasp-and-insert skill was executed across 15 different spatial configurations where the measurement station position and orientation varied systematically (see \cref{fig:skill_comb_positions}).
Configurations included standard positions (0$^\circ$) and poses where the measurement station was rotated to 180$^\circ$ relative to training demonstrations.
For each configuration, the composed skill was executed once.
Success required: (1) successful grasp of bearing ring, (2) collision-free transport to measurement station, and (3) precise insertion into measurement unit.

\textbf{Results: 73.3\% (11/15 trials)}: Testing with lateral position variations shows successful insertion at multiple measurement unit locations across the workspace, validating that the frame-relative encoding enables position-invariant execution: the robot adapts its absolute motion to new measurement unit locations without retraining.
All 4 failures in the orientation experiment occurred at positions where the measurement station was rotated 180 degrees relative to training demonstrations.
This aligns with results from \cite{Knauer2025}.
At these configurations, the required insertion motion is fundamentally opposite to the learned trajectories.
Additional demonstrations in rotated configurations would not help, as they would introduce contradictory training data conflicting with existing demonstrations, as the TP-KMP would attempt to average fundamentally different motion directions \cite{Paraschos2013, Saveriano2023}.

\subsection{Additional Evaluation Figures}\label{sec:appendix_eval_figures}
This section collects additional evaluation figures referenced in the main text.

\begin{figure}[tb]
\centering
\includegraphics[width=0.55\linewidth]{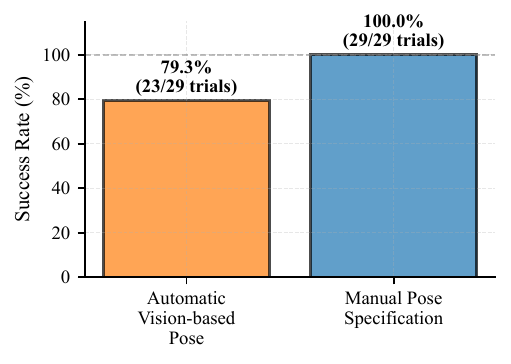}
\caption{Pose generalization robustness. Comparison of automatic vision-based pose estimation (79.3\% success) versus manual pose specification (100\% success) demonstrates that perception accuracy is the bottleneck for fully autonomous execution.}
\label{fig:pose_gen}
\end{figure}

\begin{figure}[!tb]
	\centering
	\includegraphics[width=\linewidth]{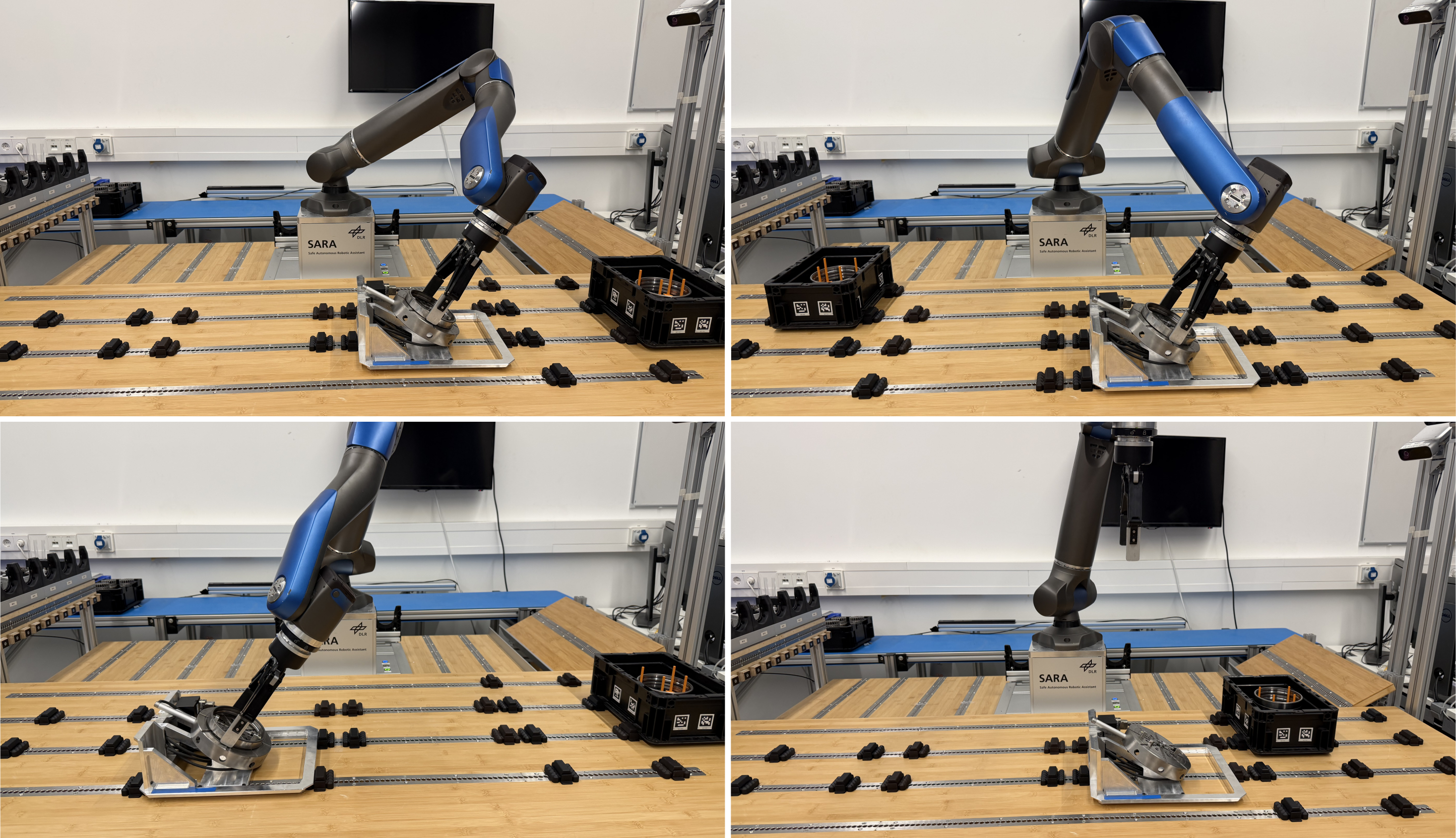}
	\caption{Examples of the skill combination robustness evaluation showing different box and measurement station positions we evaluated on.}
	\label{fig:eval_skill_combination}
\end{figure}

\begin{figure}[!tb]
	\centering
	\includegraphics[width=\linewidth]{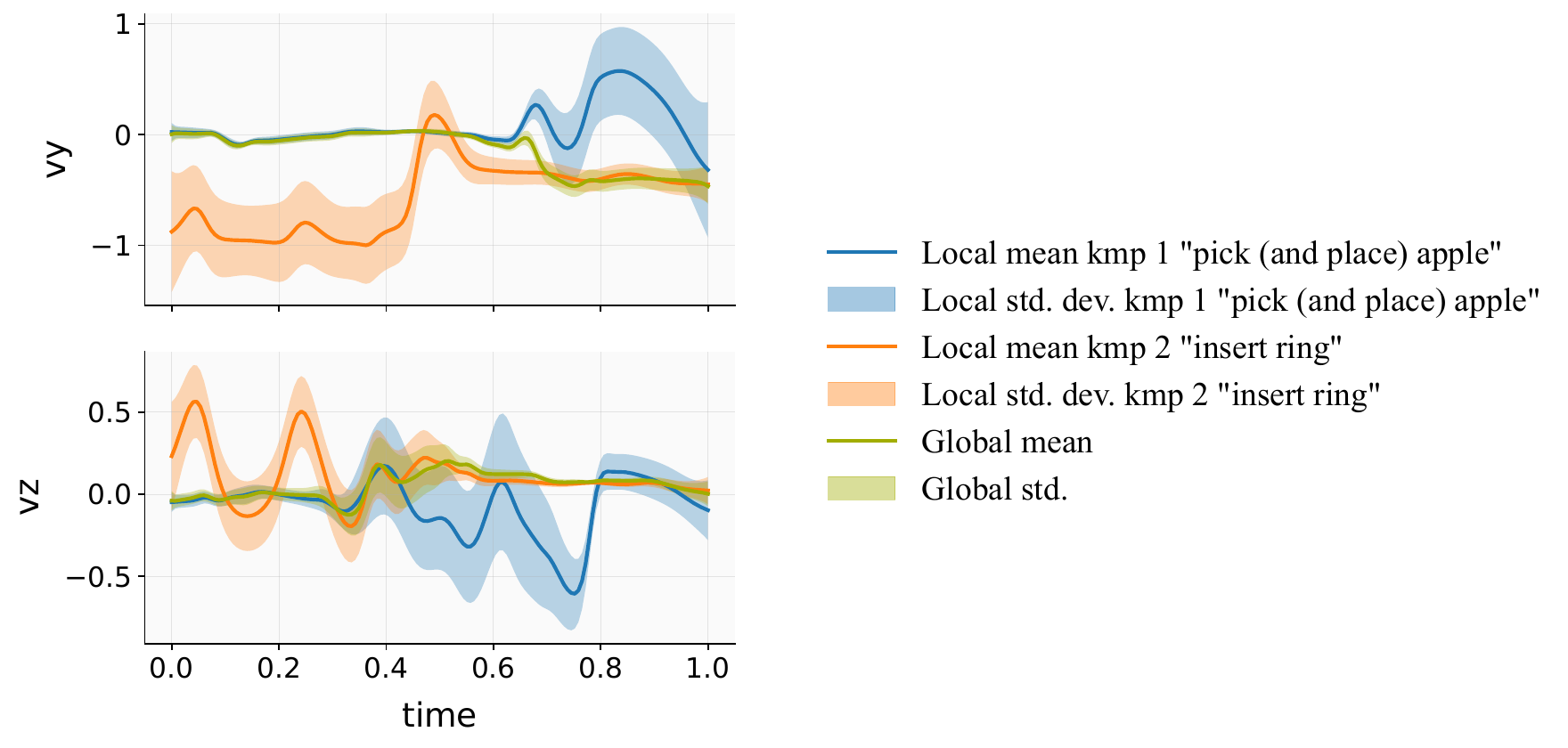}
	\caption{Trajectory fusion of grasp and insert skills via TP-KMP covariance-weighted composition. Demonstrations are temporally aligned using dynamic time warping prior to training. Shaded regions indicate frame dominance based on covariance structure.}
	\label{fig:eval_skill_combination_traj}
\end{figure}

\subsection{Covariance Structure Analysis for Active Learning Composition}\label{sec:appendix_covariance_analysis}

Analyzing the covariance structures of the composed grasp-and-insert skill: the grasp skill exhibits low covariance in the grasp phase, and high covariance in the place motion.
The \textit{insert ring} skill exhibits high covariance in the approach and low covariance during insertion (precision required).
The first frame of grasp-place is compatible with the \textit{insert ring} frame because their confidence regions do not overlap during the grasp motion.
\Cref{fig:eval_skill_combination_traj} shows the detailed trajectory analysis of this composition, visualizing how the covariance-weighted fusion enables smooth motion blending between the grasp and insertion phases.

\subsection{Limitations}\label{sec:appendix_limitations}

\textbf{Composition constraints:} Skill composition requires satisfying the compatibility constraint (\cref{eq:compatibility_constraint}) and is currently validated for pairwise combinations. Not all skill pairs are compatible; conflicting confidence regions produce unintended motion blending.
In principle, a composed TP-KMP (with $P = 2$) is itself a standard TP-KMP, so one of its local KMPs could be selected for further composition with a third skill, enabling incremental multi-skill chaining. However, the compatibility constraint becomes progressively harder to satisfy with depth.
\textbf{Static scene assumption:} The framework assumes object poses remain constant during execution. Dynamic scenes would require real-time pose re-estimation and trajectory replanning.
\textbf{Perception dependency:} Results depend on accurate 6D pose estimation; perception failures impact execution success (79.3\% vs 100\% with manual poses). The current setup uses a single static RGB-D camera, which contributes to occlusion-related failures. Multi-view estimation would likely reduce this gap.
\textbf{Comparison with VLAs:} End-to-end VLA approaches offer broader semantic generalization and can handle novel task descriptions without explicit skill definitions. Furthermore, VLAs handle perception end-to-end and can potentially generalize to novel object categories zero-shot, whereas our modular approach requires training the perception pipeline (object detection and 6D pose estimation) for each new object class, a cost not captured by the 2--5 demonstration count for motion learning alone. Our approach trades this flexibility for data efficiency in the motion learning component and interpretability, targeting specialized industrial settings where the set of manipulated objects is known and precision is prioritized.

\end{document}